\documentclass[11pt]{article}
\usepackage{arxiv}
\usepackage{amsmath}
\DeclareUnicodeCharacter{00A7}{\S}
\DeclareUnicodeCharacter{00B0}{\textdegree}
\DeclareUnicodeCharacter{00B1}{\ensuremath{\pm}}
\DeclareUnicodeCharacter{00D7}{\ensuremath{\times}}
\DeclareUnicodeCharacter{0394}{\ensuremath{\Delta}}
\DeclareUnicodeCharacter{03C3}{\ensuremath{\sigma}}
\DeclareUnicodeCharacter{2026}{\ldots}
\DeclareUnicodeCharacter{2192}{\ensuremath{\rightarrow}}
\DeclareUnicodeCharacter{2212}{\ensuremath{-}}
\DeclareUnicodeCharacter{2248}{\ensuremath{\approx}}
\DeclareUnicodeCharacter{226A}{\ensuremath{\ll}}
\DeclareUnicodeCharacter{2265}{\ensuremath{\ge}}
\usepackage[table]{xcolor}
\definecolor{TietTableStripe}{HTML}{F3F6FA}
\definecolor{TietTableRule}{HTML}{CBD5E1}
\usepackage{color}
\usepackage{fancyvrb}

\DefineVerbatimEnvironment{Highlighting}{Verbatim}{commandchars=\\\{\}}
\usepackage{framed}
\definecolor{shadecolor}{RGB}{241,243,245}
\newenvironment{Shaded}{\begin{snugshade}}{\end{snugshade}}

\newcommand{\DataTypeTok}[1]{\textcolor[rgb]{0.68,0.00,0.00}{#1}}

\newcommand{\FunctionTok}[1]{\textcolor[rgb]{0.28,0.35,0.67}{#1}}

\newcommand{\NormalTok}[1]{\textcolor[rgb]{0.00,0.23,0.31}{#1}}

\newcommand{\OtherTok}[1]{\textcolor[rgb]{0.00,0.23,0.31}{#1}}

\newcommand{\StringTok}[1]{\textcolor[rgb]{0.13,0.47,0.30}{#1}}

\usepackage{graphicx}
\usepackage{subcaption}
\makeatletter
\@ifpackageloaded{caption}{}{\usepackage{caption}}
\AtBeginDocument{%
\ifdefined\contentsname
  \renewcommand*\contentsname{Table of contents}
\else
  \newcommand\contentsname{Table of contents}
\fi
\ifdefined\listfigurename
  \renewcommand*\listfigurename{List of Figures}
\else
  \newcommand\listfigurename{List of Figures}
\fi
\ifdefined\listtablename
  \renewcommand*\listtablename{List of Tables}
\else
  \newcommand\listtablename{List of Tables}
\fi
\ifdefined\figurename
  \renewcommand*\figurename{Figure}
\else
  \newcommand\figurename{Figure}
\fi
\ifdefined\tablename
  \renewcommand*\tablename{Table}
\else
  \newcommand\tablename{Table}
\fi
}
\@ifpackageloaded{float}{}{\usepackage{float}}
\floatstyle{ruled}
\@ifundefined{c@chapter}{\newfloat{codelisting}{h}{lop}}{\newfloat{codelisting}{h}{lop}[chapter]}
\floatname{codelisting}{Listing}

\makeatother
\makeatletter
\@ifpackageloaded{caption}{}{\usepackage{caption}}
\@ifpackageloaded{subcaption}{}{\usepackage{subcaption}}
\makeatother

\title{Small data but Specific: Task-Aligned Embeddings for Clinical
Code Retrieval}

\author{%
David Rey-Blanco%
\thanks{\texttt{david.rey@tiet.ai}}%
 \qquad %
Roberto Cruz%
\thanks{\texttt{roberto.cruz@tiet.ai}}%
\\[0.4em]
\small TietAI%
}

\date{2026-04-10}

\begin{document}
\maketitle

\begin{abstract}
\noindent Clinical coding can be framed as a fine-grained semantic
retrieval task in which short clinical mentions must be matched to the
correct ICD-10/CIE-10 code despite subtle distinctions in severity,
anatomy, temporality, or etiology. Existing sentence-embedding models
are poorly adapted to this setting, particularly for non-English
clinical text, while manually annotated training data remains scarce and
expensive. This work studies whether compact, task-specific retrievers
can be improved through high-quality synthetic supervision generated
with large language models.

We use a frontier Large Language Model to construct multilingual
training corpora grounded in the ICD-10 hierarchy. A Spanish biomedical
bi-encoder is fine-tuned on 19,502 synthetic examples across six
languages, while a cross-encoder reranker is trained on 10,628 listwise
groups with clinically plausible hard negatives. The resulting system is
evaluated on CodiESP v4 and DISTEMIST against BM25 and several widely
used sentence-transformer baselines.

The proposed retriever achieves F1 = 0.709 and MAP@10 = 0.747 on
CodiESP, and F1 = 0.776 and MAP@10 = 0.812 on DISTEMIST, outperforming
all public baselines by a large margin. Results indicate that retrieval
quality depends more on task-aligned supervision and clinically
meaningful negatives than on model scale alone. Our findings suggest
that carefully curated synthetic datasets can enable compact and
effective clinical semantic search systems without requiring web-scale
training corpora.
\end{abstract}

\bigskip

\section{Introduction}\label{introduction}

Clinical coding is a high-stakes text-to-code retrieval problem. Given a
short clinical mention, complaint, or fragment of a medical note, the
system must retrieve the correct entry from a controlled terminology
such as ICD-10-CM or its Spanish adaptation, CIE-10. This is harder than
ordinary semantic search because adjacent codes often describe
clinically similar conditions that differ by a qualifier such as
organism, laterality, severity, episode type, or anatomical location. A
useful model must therefore retrieve the right clinical variant, not
only the right disease family.

Modern dense retrieval usually addresses this kind of problem with two
stages. A bi-encoder first projects queries and candidate descriptions
into a shared vector space and retrieves a shortlist efficiently
\citep{reimers2019sbert, karpukhin2020dpr}. A cross-encoder then jointly
reads the query and each candidate in the shortlist to produce a more
precise relevance score
\citep{nogueira2019passagererank, hofstaetter2020intradocument}. This
recall-then-rerank architecture is common in retrieval-augmented
generation \citep{lewis2020rag, xiong2024medrag, ragmedical2025} and in
automatic ICD coding
\citep{mullenbach2018caml, ji2021icdcontrastive, yuan2022icdcoder}.

The difficulty is that most public embedding models and biomedical NLP
resources are still centred on English. BioBERT \citep{lee2020biobert},
ClinicalBERT \citep{alsentzer2019clinicalbert}, and related models are
strong on English EHRs and biomedical benchmarks such as PubMedQA
\citep{jin2019pubmedqa}, but Spanish, Catalan, Italian, Portuguese, and
other clinical languages are under-represented
\citep{joshi2020stateandfate}. Multilingual encoders help
\citep{conneau2020xlmr, pires2019multilingualbert, reimers2020multilingualsbert},
but clinical retrieval remains fragile when abbreviations, morphology,
surface forms, and coding conventions vary across languages
\citep{neveol2018clinicalmultilang}.

\subsection{The limitations of clinical coding
models}\label{the-limitations-of-clinical-coding-models}

The central issue is not only the lack of model capacity.
General-purpose encoders are usually trained to capture broad semantic
relatedness, whereas clinical coding requires fine-grained equivalence.
For example, \emph{``Brucelosis debida a Brucella suis''} and
\emph{``Brucelosis debida a Brucella melitensis''} are semantically
close but not interchangeable as codes. The same tension appears across
thousands of CIE-10 descriptions: the nearest clinical neighbour is
often the wrong answer.

Four practical limitations follow from this setting. First, tokenisation
and pretraining corpora under-cover non-English clinical morphology,
which can flatten important lexical distinctions. Second, biomedical
pretraining does not necessarily transfer if it was learned in another
language or on a different clinical distribution
\citep{lee2020biobert, alsentzer2019clinicalbert, carrino2022bsc}.
Third, cross-lingual transfer degrades in dense retrieval because small
representation shifts can change the top-1 result
\citep{lauscher2020zerohero, wu2019crosslingualeval}. Fourth, two-stage
systems can be misaligned: a reranker cannot recover the correct code if
the bi-encoder never retrieves it, and a high-recall bi-encoder is
insufficient if it cannot order clinically adjacent candidates.

The data problem is equally important. Labelled clinical data is
expensive because annotation requires both medical and linguistic
expertise. In multilingual coding, the data must also match the local
coding vocabulary and the intended level of granularity. A corpus
labelled at chapter level does not teach full-code distinctions, and an
English ICD corpus does not necessarily contain the Spanish, Catalan, or
Portuguese expressions that a local coding system must handle.

\subsection{Proposal, results, and
contributions}\label{proposal-results-and-contributions}

We study whether LLM-generated synthetic supervision can make domain-
and language-specific clinical retrievers practical under these data
constraints. Rather than using synthetic data only as generic
augmentation, we use an LLM as a tutor for a controlled label space:
given a code, definition, neighbouring codes, and target language, the
generator creates clinical mentions, paraphrases, and hard negatives
that force the model to learn the decision boundary. This follows recent
work on LLM-generated instruction and retrieval data
\citep{wang2023selfinstruct, bonifacio2022inpars, wang2024promptagator}
and on clinical synthetic text generated from domain knowledge or labels
\citep{peng2023gatortrongpt, xu2024clingen, li2023twodirections, kumichev2024medsyn}.

Concretely, we train a two-stage retriever for Spanish CIE-10/ICD-10
search. The bi-encoder is fine-tuned from
\emph{PlanTL-GOB-ES/bsc-bio-ehr-es} \citep{carrino2022bsc} on about
19,500 multilingual synthetic pairs covering English, Spanish, Catalan,
Italian, Portuguese, and French. The cross-encoder reranker is trained
on roughly 10,600 listwise groups with clinically plausible hard
negatives. We evaluate the resulting system on two held-out Spanish
clinical coding datasets, CodiESP v4 and DISTEMIST, at full-code,
category, and chapter resolution.

The main result is that targeted synthetic supervision and reranking
produce a large improvement over public baselines. On CodiESP, the
complete system reaches 70.9\% top-1 exact accuracy, improving by 48.4
percentage points over the best public encoder tested, MiniLM-L6-v2, and
by 51.6 points over the in-domain English S-BioBERT baseline. On
DISTEMIST, it reaches 77.6\% top-1 exact accuracy. The results also show
that larger embeddings and biomedical pretraining alone are not enough:
MPNet-v2 underperforms MiniLM-L6-v2, while S-BioBERT underperforms both
despite domain-specific pretraining.

The paper makes three contributions:

\begin{enumerate}
\def\labelenumi{\arabic{enumi}.}
\tightlist
\item
  It documents a reproducible recipe for training a two-stage ICD
  retriever from multilingual LLM-generated supervision.
\item
  It shows that language coverage and task-aligned synthetic curation
  can matter more than generic biomedical pretraining for Spanish
  clinical code retrieval.
\item
  It quantifies the role of cross-encoder reranking in resolving
  qualifier-level distinctions that are common in ICD-10-CM and
  CIE-10-ES.
\end{enumerate}

Synthetic data is not treated here as a substitute for real clinical
evaluation, thus generated examples can be too regular, biased toward
the generator's preferred phrasing, or detached from the noise of real
notes \citep{long2024synthsurvey, shumailov2023modelcollapse}. For that
reason, we use synthetic data only for supervision and evaluate
exclusively on held-out human clinical coding datasets (using
human-anotated sets only for this task).

This paper is structured as follows. The methodology section describes
the synthetic data generation process, the bi-encoder, and the
cross-encoder. The experimental design section defines the CIE-10
hierarchy, the search backend, the baseline models, the evaluation
datasets, and the metrics. The results section reports performance on
CodiESP and DISTEMIST. The discussion compares the findings with public
benchmarks and related ICD coding literature, and the final sections
summarize limitations and future work.

\section{Methodology}\label{methodology}

\subsection{System overview}\label{system-overview}

We train a two-stage neural retrieval architecture for ICD code search
following the retrieve-and-rerank paradigm commonly used in dense
information retrieval systems
\citep{reimers2019sbert, karpukhin2020dpr}. In the first stage, a
bi-encoder retrieves the top-k candidate ICD descriptions using cosine
similarity in a shared embedding space. In the second stage, a
cross-encoder jointly models the query and each retrieved candidate to
rerank results by semantic relevance \citep{nogueira2019passagererank}.

The bi-encoder is initialised from \emph{PlanTL-GOB-ES/bsc-bio-ehr-es}
\citep{carrino2022bsc} and fine-tuned with
\emph{MultipleNegativesRankingLoss} \citep{henderson2017mnr} using
multilingual synthetic positive pairs and in-batch negatives, while the
reranker is optimised with a listwise softmax cross-entropy objective
\citep{cao2007learning} over one positive and up to nine hard negatives
per query.

\begin{figure}[H]

\centering{

\pandocbounded{\includegraphics[keepaspectratio]{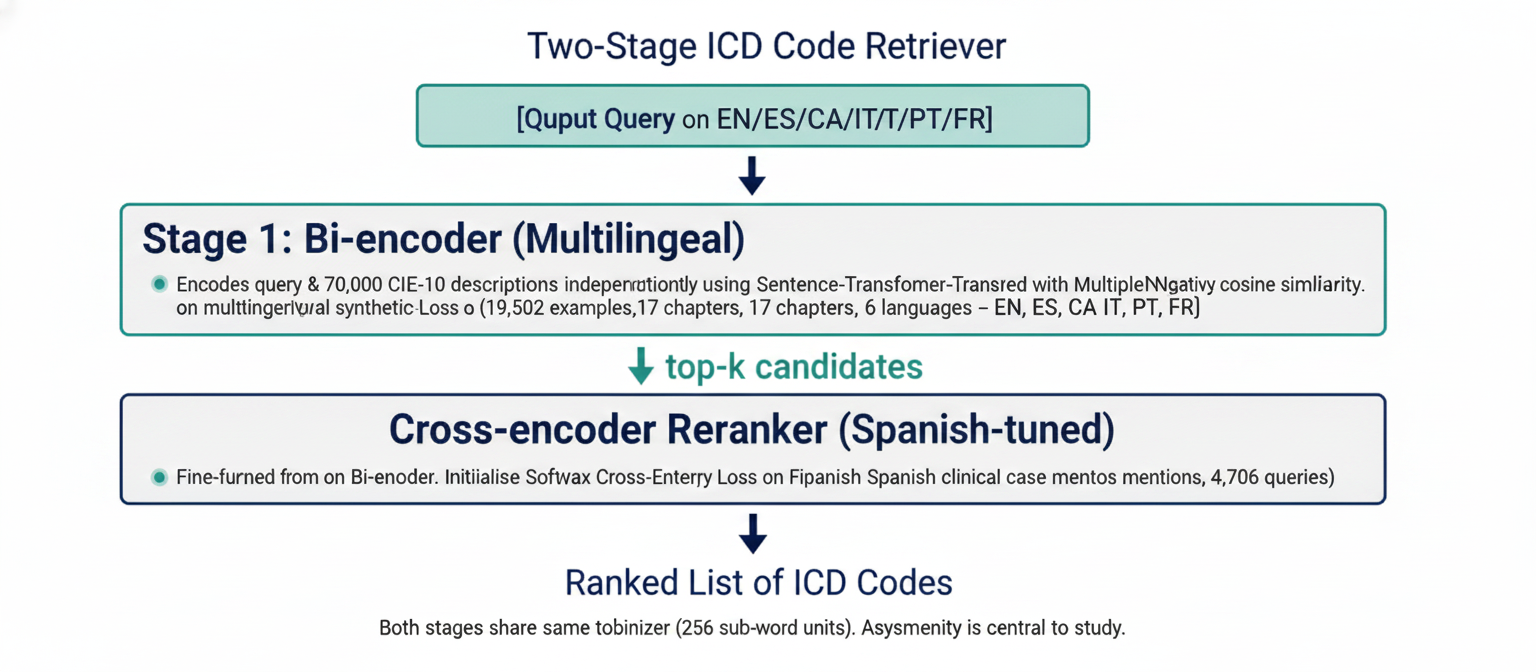}}

}

\caption{\label{fig-pipeline}Two-stage retriever: a multilingual
bi-encoder feeds a Spanish-tuned cross-encoder reranker. The asymmetry
between the two stages is the central object of this study. Source:
internal}

\end{figure}

The pipeline is summarised in Figure~\ref{fig-pipeline}. Both stages
share the same backbone tokenizer and maximum sequence length (256
sub-word units), making the reranker initialisation trivial.

\subsection{Synthetic data generation for
training}\label{synthetic-data-generation-for-training}

We use Google DeepMind Gemini 2.5 Flash Pro \citep{gemini25pro} as a
data factory to generate two dataset families grounded on the ICD-10
chapter hierarchy:

\begin{enumerate}
\def\labelenumi{\arabic{enumi}.}
\tightlist
\item
  Dataset A (bi-encoder training). This set is designed to train dense
  retrieval models for multilingual clinical semantic search. Each
  example contains a free-text clinical query, two positives
  (paraphrases or canonical ICD-10 descriptions), two hard negatives
  sampled from semantically adjacent codes within the same chapter, and
  two soft negatives sampled from unrelated chapters. Data generation is
  parameterised by ICD-10 chapter and target language (we use the
  chapter as a seed). We generate examples in English, Spanish, Catalan,
  Italian, Portuguese, and French. After deduplication and quality
  filtering, the final corpus contains 19,502 examples across 17 ICD-10
  chapters and 6 languages (see Table~\ref{tbl-bi-data-stats}), and a
  set of examples generated without a chapter random seed (labeled as
  \emph{Ungrouped} in the table).
\item
  Dataset B (cross-encoder training). Following the CodiEsp track of
  CLEF eHealth \citep{miranda2020codiesp}, we anchor cross-encoder
  training on real Spanish clinical case mentions. For each annotated
  mention we collect one positive CIE-10 description and up to nine hard
  negatives sampled by lexical and embedding adjacency. The final
  training file contains 10,628 listwise groups over 4,706 unique
  queries in Spanish.
\end{enumerate}

Dataset A was intentionally designed as a compact but high-quality
synthetic corpus. Rather than relying on web-scale weak supervision, we
focus on clinically grounded examples with difficult in-chapter
negatives that force the bi-encoder to learn fine-grained semantic
distinctions between related diagnoses. This design follows the
hypothesis that, for specialised medical retrieval tasks, carefully
curated synthetic supervision can compensate for limited dataset size. A
central aspect of the dataset construction is the balance between hard
and soft negatives. Hard negatives expose the model to clinically
adjacent concepts that may differ only in severity, anatomical location,
temporality, or etiology, encouraging the encoder to learn subtle
decision boundaries within the ICD hierarchy. In contrast, soft
negatives sampled from unrelated chapters prevent representation
collapse and preserve global semantic separation across the embedding
space. This balance is particularly important in clinical coding, where
code assignment is often ambiguous and partially subjective, and where
many retrieval errors arise not from lexical mismatch but from confusion
between semantically neighbouring diagnoses. To improve consistency and
reduce noise, all generated samples undergo a multi-stage curation
pipeline including rule-based validation, semantic deduplication,
language consistency checks, and manual spot inspection. The resulting
dataset is substantially smaller than large-scale sentence embedding
corpora used in general-domain retrieval models, which are often trained
on hundreds of millions of text pairs \citep{deka2021unsupervised}, yet
it remains sufficiently expressive for domain-specific semantic search.

Dataset A is multilingual, whereas Dataset B is anchored in Spanish
CodiEsp-derived supervision. This setup reflects realistic annotation
availability constraints for Spanish clinical AI teams.

\begin{longtable}[]{@{}
  >{\raggedright\arraybackslash}p{(\linewidth - 2\tabcolsep) * \real{0.8000}}
  >{\raggedleft\arraybackslash}p{(\linewidth - 2\tabcolsep) * \real{0.1000}}@{}}
\caption{Bi-encoder synthetic corpus by ICD-10 chapter (top 10 of 17
shown). The \emph{ungrouped} row corresponds to the CodiEsp-derived seed
examples used for back-translation and
warm-up.}\label{tbl-bi-data-stats}\tabularnewline
\toprule\noalign{}
\begin{minipage}[b]{\linewidth}\raggedright
Chapter
\end{minipage} & \begin{minipage}[b]{\linewidth}\raggedleft
N
\end{minipage} \\
\midrule\noalign{}
\endfirsthead
\toprule\noalign{}
\begin{minipage}[b]{\linewidth}\raggedright
Chapter
\end{minipage} & \begin{minipage}[b]{\linewidth}\raggedleft
N
\end{minipage} \\
\midrule\noalign{}
\endhead
\bottomrule\noalign{}
\endlastfoot
Congenital malformations, deformations and chromosomal abnormalities (Q)
& 1,139 \\
Endocrine, nutritional and metabolic diseases (E) & 1,123 \\
Diseases of the circulatory system (I) & 1,119 \\
Certain infectious and parasitic diseases (A-B) & 1,115 \\
Mental, behavioral and neurodevelopmental disorders (F) & 1,109 \\
Diseases of the musculoskeletal system and connective tissue (M) &
1,092 \\
Diseases of the respiratory system (J) & 1,085 \\
Neoplasms (C-D) & 1,084 \\
Diseases of the skin and subcutaneous tissue (L) & 1,084 \\
Diseases of the nervous system (G) & 1,078 \\
\emph{Ungrouped / random seed} & 2,906 \\
\textbf{Total (all 17 chapters)} & \textbf{19,502} \\
\end{longtable}

Representative JSON rows for the bi-encoder and cross-encoder corpora
are provided in Section~\ref{sec-annex-dataset-examples}.

\subsection{Bi-encoder training}\label{bi-encoder-training}

We fine-tune \emph{PlanTL-GOB-ES/bsc-bio-ehr-es} BERT based model with
\texttt{MultipleNegativesRankingLoss} over (query, positive) pairs
flattened from Dataset A. In-batch negatives provide the contrastive
signal. For a batch of \(B\) query-positive pairs
\(\{(q_i, p_i)\}_{i=1}^{B}\), the loss for query \(q_i\) is defined as:

\[
\mathcal{L}_i =
-\log
\frac{
\exp \left( s(q_i, p_i) / \tau \right)
}{
\sum_{j=1}^{B}
\exp \left( s(q_i, p_j) / \tau \right)
}
\]

where \(s(q_i, p_j)\) is the similarity score, typically cosine
similarity or dot product between the bi-encoder embeddings, and
\(\tau\) is the temperature parameter. The batch loss is:

\[
\mathcal{L} =
\frac{1}{B}
\sum_{i=1}^{B}
\mathcal{L}_i
\]

Hyperparameters follow standard sentence-transformer practice and were
not extensively tuned: batch size 32, three epochs, AdamW with 10\%
linear warmup, max sequence length 256 and mean-pooling.

Although the supervised signal contains only positive pairs, each batch
draws texts from all six languages. The in-batch contrastive loss can
therefore penalise cross-lingual confusions without explicit translation
pairs \citep{reimers2020multilingualsbert}.

Information-retrieval evaluation during training is performed on a 10\%
held-out slice of \emph{rows} (not flattened pairs) with up to 64
negatives per query drawn from the union of hard and soft negatives and
a global pool fallback. We track Recall@\(k\), MRR and nDCG@\(k\).

\subsection{Cross-encoder training}\label{cross-encoder-training}

The cross-encoder is a \emph{CrossEncoder} model with one regression
head, initialised from the bi-encoder checkpoint. Each batch contains 8
grouped examples; each group is one query plus 1 positive and up to 9
hard negatives sampled from the listwise pool. We compute logits for the
\(1+9\) pairs and apply softmax cross-entropy with the positive at index
0 \citep{nogueira2019passagererank}. Training uses AdamW at a learning
rate of 2 × 10\textsuperscript{-5} with 10\% warmup for 5 epochs,
gradient clipping at norm 1.0.

The cross-encoder is trained exclusively on Spanish CodiEsp-derived
data, although the bi-encoder sees six languages. This reflects the
higher cost of obtaining listwise supervision in minority languages and
allows us to measure the effect of this stage-level mismatch.

\subsection{Experimental design}\label{sec-experimental-design}

We evaluate the TietAI bi-encoder and cross-encoder against
representative lexical and sentence-transformer baselines on two
independent Spanish clinical-coding corpora.

\subsubsection{Coding: CIE-10-ES and the ICD-10
hierarchy}\label{coding-cie-10-es-and-the-icd-10-hierarchy}

The International Classification of Diseases, 10th revision (ICD-10) is
a controlled vocabulary maintained by the World Health Organisation in
which every clinical concept --- diagnosis, sign, external cause, social
circumstance --- is mapped to an alphanumeric code drawn from a strict
hierarchy. Codes are organised into 22 chapters (e.g.~\emph{I. Certain
infectious and parasitic diseases} covering \emph{A00--B99}, \emph{IX.
Diseases of the circulatory system} covering \emph{I00--I99}). Within a
chapter each code opens with a three-character category that names the
disease family --- for example \emph{K35} for \emph{acute appendicitis}
--- and then resolves to a finer-grained code by appending up to four
additional characters that specify laterality, severity, episode of care
or anatomical site, e.g. \emph{K35.20} \emph{acute appendicitis with
generalized peritonitis, without abscess}. Two codes that share their
first three characters therefore denote clinically related conditions,
while two codes that share only their first character merely belong to
the same chapter.

CIE-10-ES is the Spanish national adaptation, currently in its 6th
edition (Ministerio de Sanidad, Servicios Sociales e Igualdad), and is
the version that Spanish hospitals are legally required to use for
diagnostic and procedural reporting. CIE-10-ES inherits the chapter /
category / full-code hierarchy of ICD-10 unchanged, but the textual
descriptions are translated and, in some cases, refined to reflect
Spanish clinical practice (e.g.~\emph{Brucelosis debida a Brucella suis}
vs. ICD-10's \emph{Brucellosis due to Brucella suis}). This three-level
structure (chapter → category → full code) is the granularity at which
we evaluate every model: a hit at \emph{chapter} level means the model
placed the query in the right disease family, a hit at \emph{category}
level means it identified the right disease, and a hit at \emph{full
code} level means it identified the right clinical variant.

\subsubsection{Models compared}\label{models-compared}

We benchmark six retrieval strategies: the two-stage TietAI pipeline, a
lexical baseline, and three off-the-shelf sentence-transformer encoders.

\begin{itemize}
\item
  \emph{TietAI Bi-Encoder (dense, 768 d)}. The Spanish biomedical
  encoder described in the Methodology section. It is the \emph{Stage 1}
  model and the baseline against which the reranker's contribution is
  measured.
\item
  \emph{TietAI Cross-Encoder.} Reranks the top-\(k\) candidates returned
  by the bi-encoder. It reads each (query, candidate) pair jointly and
  produces a relevance score for sibling CIE-10 codes that may have
  near-identical bi-encoder cosine distances.
\item
  \emph{BM25 over full-text search.} A classical sparse baseline over
  the CIE-10-ES descriptions. It measures how much retrieval quality can
  be explained by lexical overlap.
\item
  \emph{Sentence-Transformers MiniLM-L6-v2 (384 d)}
  \citep{reimers2019sbert}. A general-purpose distilled encoder trained
  on web/NLI pairs, included as a generalistic control.
\item
  \emph{Sentence-Transformers BioBERT-ST (768 d)}
  (\emph{pritamdeka/S-BioBert-snli-multinli-stsb})
  \citep{deka2021unsupervised}, \citep{lee2020biobert}. A biomedical
  sentence encoder fine-tuned on English EHR-style data, included to
  test whether English biomedical pretraining transfers to Spanish
  CIE-10-ES retrieval.
\item
  \emph{Sentence-Transformers MPNet-base-v2 (768 d)}
  \citep{song2020mpnet}. A larger general-purpose encoder included to
  evaluate whether higher embedding dimensionality improves retrieval
  quality.
\end{itemize}

\subsubsection{Evaluation datasets}\label{evaluation-datasets}

We use two independent CIE-10-ES corpora. CodiESP v4
\citep{miranda2020codiesp} is the development set of the CLEF eHealth
2020 shared task on automatic clinical coding in Spanish. It provides
3,664 mention-to-code pairs annotated from 250 case reports
(\emph{testD.tsv} for diagnoses, \emph{testP.tsv} for procedures and
\emph{testX.tsv} for the textual spans). For each row we treat the
mention text as the query and the gold CIE-10-ES code as the relevance
judgment. CodiESP is our \emph{primary} benchmark and we evaluate every
model on it.

DISTEMIST \citep{miranda2022distemist} is a complementary corpus of
disease mentions extracted from Spanish clinical cases, released as part
of the BioASQ track and distributed on the BigBio repository. We use the
\emph{subtrack 2 linking} split with cross-mappings (1,224 disease
mentions with a CIE-10 code). DISTEMIST is used as a \emph{secondary}
benchmark whose role is narrower: it tests whether the conclusions drawn
on CodiESP extrapolate to another Spanish clinical corpus with a
different provenance and a different distribution of disease mentions.
For this reason we report DISTEMIST only for the TietAI stack and the
BM25 baseline.

\subsubsection{Search backend and
metrics}\label{search-backend-and-metrics}

All dense retrievers are served from a Qdrant vector database (cosine
similarity, HNSW index) populated with the CIE-10-ES 6th edition
catalogue. The same code descriptions are indexed in Postgres for the
BM25 full-text channel, used as the standalone lexical baseline. For
each query we retrieve the top-10 candidates and report exact-code
accuracy at \(k \in \{1, 3, 5, 10\}\), together with the corresponding
scores at the \emph{category} (first three characters) and
\emph{chapter} level. Top-1 accuracy at each of the three hierarchical
resolutions is summarised separately to make the calibration / coverage
trade-off explicit.

\section{Results}\label{results}

\subsection{Evaluation metrics}\label{sec-metrics}

Automatic CIE-10-ES coding is a \emph{multi-label retrieval} problem
with a strongly imbalanced label space (tens of thousands of candidate
codes, typically a handful of correct codes per document). Each metric
we report answers a slightly different question, and the right one
depends on the downstream use of the retriever. We list them below in
decreasing order of operational relevance for clinical coding, and we
report all of them --- but the headline figures in
Section~\ref{sec-codiesp-results} and
Section~\ref{sec-distemist-results} are F1 and MAP@10, which we consider
the two most informative summaries for this task.

\subsubsection*{Mean Average Precision at K
(MAP@K)}\label{mean-average-precision-at-k-mapk}
\addcontentsline{toc}{subsubsection}{Mean Average Precision at K
(MAP@K)}

For a query \(q\) with relevant set \(\mathcal{R}_q\) and a ranked list
of the top \(K\) candidates, the Average Precision at K (MAP@K) is the
mean of AP@K over the evaluation set. MAP rewards \emph{ranking
quality}: a system that places the correct code at rank 1 scores higher
than one that places it at rank 5, even if both retrieve it within the
top 10.

\[
\text{AP@K}(q) \;=\; \frac{1}{\min(|\mathcal{R}_q|, K)} \sum_{k=1}^{K} \text{Precision@}k(q) \cdot \mathbf{1}\!\left[\text{item}_k \in \mathcal{R}_q\right]
\]

This is the metric of choice when the retriever feeds a human coder or
an agentic reranker that inspects the top shortlist: clinically, the
right code should \emph{appear early}. The CodiESP shared task uses MAP
as its primary diagnosis metric \citep{miranda2020codiesp}, and we
follow the same convention.

\subsubsection*{F1 score}\label{f1-score}
\addcontentsline{toc}{subsubsection}{F1 score}

F1 is the harmonic mean of precision and recall and therefore penalises
either failure mode (too many false positives, or too many missed
diagnoses) symmetrically.

\[
F_1 \;=\; \frac{2 \cdot \text{Precision} \cdot \text{Recall}}{\text{Precision} + \text{Recall}}
\]

It is the most useful \emph{single} summary of a coding system because
in this domain both errors carry cost: false positives create billing
and epidemiological noise, false negatives result in incomplete medical
records and under-coded morbidity.

\subsubsection*{Recall}\label{recall}
\addcontentsline{toc}{subsubsection}{Recall}

Recall measures how many of the truly relevant codes the system
recovers.

\[
\text{Recall} \;=\; \frac{\text{TP}}{\text{TP} + \text{FN}}
\]

In clinical settings missing a diagnosis (e.g.~sepsis, a diabetes
complication, an oncological staging code) tends to be more costly than
adding a spurious one, so recall has a natural priority. Reported in
isolation it is, however, easy to game: a system that returns the entire
CIE-10-ES catalogue for every query scores recall \(=1\) and is useless.

\subsubsection*{Precision}\label{precision}
\addcontentsline{toc}{subsubsection}{Precision}

Precision measures how many of the system's predictions are correct.

\[
\text{Precision} \;=\; \frac{\text{TP}}{\text{TP} + \text{FP}}
\]

It matters because each spurious code propagates into billing, reporting
and EHR analytics. Like recall, it is easy to game in the opposite
direction: an extremely conservative system that predicts a code only
when it is near-certain can reach very high precision while being
clinically useless. The CodiESP shared task contains explicit examples
of this trade-off (see Section~\ref{sec-litcomp}).

\subsubsection*{Accuracy}\label{accuracy}
\addcontentsline{toc}{subsubsection}{Accuracy}

\[
\text{Accuracy} \;=\; \frac{\text{TP} + \text{TN}}{\text{TP} + \text{TN} + \text{FP} + \text{FN}}
\]

Accuracy is the least informative metric for ICD-style retrieval. In a
multi-label setting with thousands of candidate codes and only a few
correct ones per document, a system that predicts \emph{no code} for
every query trivially reaches accuracy close to one. We report it in the
annex for completeness but it is not our preferred summary. In the
single-gold per-mention regime we use for evaluation, accuracy coincides
with both precision and recall for the dense models (each query receives
exactly one top-1 prediction and has one gold code); the BM25 baseline
is the only system where the three metrics diverge, because for 393 of
the 3,664 CodiESP queries (≈11 \%) the lexical channel returns no
candidate at all.

In summary, MAP@10 captures \emph{how useful the system is as a ranker};
F1 captures \emph{overall quality at the top-1 decision}; recall and
precision separate the two failure modes; accuracy is reported only for
compatibility with earlier comparators.

\subsection{Results on CodiESP v4}\label{sec-codiesp-results}

\begin{figure}[H]

\centering{

\includegraphics[width=0.95\linewidth,height=\textheight,keepaspectratio]{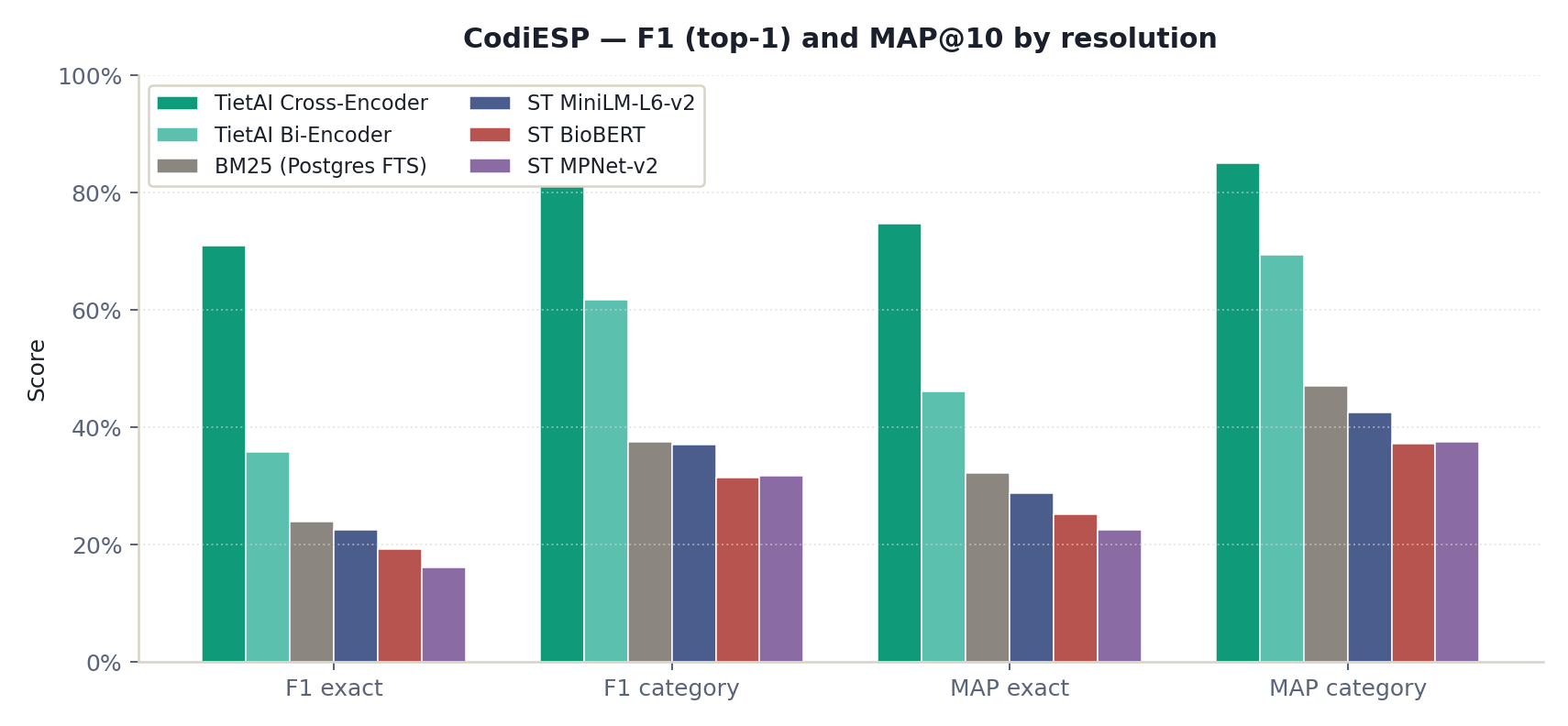}

}

\caption{\label{fig-codiesp-f1-map}CodiESP --- F1 and MAP@10 by model,
at the exact-code and three-character category levels. MAP exceeds F1
for every model, which means that for the queries where top-1 is wrong,
the correct code is still often present in the top-10 (the ranking
signal is there, the calibration is what is missing).}

\end{figure}

Table~\ref{tbl-codiesp-headline} and Figure~\ref{fig-codiesp-f1-map}
combine the four key metrics --- F1 at exact code, F1 at category,
MAP@10 at exact, and MAP@10 at category --- for all six retrievers on
the 3,664 CodiESP development mentions. The full per-resolution
breakdown (precision, recall, F1 and accuracy at chapter, category and
exact resolutions; and precision@\(k\) at \(k \in \{1,3,5,10\}\) for
both resolutions) is in Section~\ref{sec-annex-codiesp-full}.

\begin{longtable}[]{@{}
  >{\raggedright\arraybackslash}p{(\linewidth - 8\tabcolsep) * \real{0.4000}}
  >{\raggedleft\arraybackslash}p{(\linewidth - 8\tabcolsep) * \real{0.1500}}
  >{\raggedleft\arraybackslash}p{(\linewidth - 8\tabcolsep) * \real{0.1500}}
  >{\raggedleft\arraybackslash}p{(\linewidth - 8\tabcolsep) * \real{0.1500}}
  >{\raggedleft\arraybackslash}p{(\linewidth - 8\tabcolsep) * \real{0.1500}}@{}}
\caption{CodiESP v4 --- F1 at top-1 and MAP@10, at the exact-code and
three-character category levels. Best result per column in
bold.}\label{tbl-codiesp-headline}\tabularnewline
\toprule\noalign{}
\begin{minipage}[b]{\linewidth}\raggedright
Model
\end{minipage} & \begin{minipage}[b]{\linewidth}\raggedleft
F1 exact
\end{minipage} & \begin{minipage}[b]{\linewidth}\raggedleft
F1 category
\end{minipage} & \begin{minipage}[b]{\linewidth}\raggedleft
MAP@10 exact
\end{minipage} & \begin{minipage}[b]{\linewidth}\raggedleft
MAP@10 category
\end{minipage} \\
\midrule\noalign{}
\endfirsthead
\toprule\noalign{}
\begin{minipage}[b]{\linewidth}\raggedright
Model
\end{minipage} & \begin{minipage}[b]{\linewidth}\raggedleft
F1 exact
\end{minipage} & \begin{minipage}[b]{\linewidth}\raggedleft
F1 category
\end{minipage} & \begin{minipage}[b]{\linewidth}\raggedleft
MAP@10 exact
\end{minipage} & \begin{minipage}[b]{\linewidth}\raggedleft
MAP@10 category
\end{minipage} \\
\midrule\noalign{}
\endhead
\bottomrule\noalign{}
\endlastfoot
TietAI Cross-Encoder & \textbf{0.709} & \textbf{0.823} & \textbf{0.747}
& \textbf{0.851} \\
TietAI Bi-Encoder & 0.359 & 0.617 & 0.461 & 0.694 \\
BM25 (Postgres FTS) & 0.239 & 0.376 & 0.322 & 0.471 \\
ST MiniLM-L6-v2 & 0.225 & 0.371 & 0.287 & 0.426 \\
ST BioBERT & 0.193 & 0.314 & 0.252 & 0.373 \\
ST MPNet-v2 & 0.161 & 0.317 & 0.226 & 0.376 \\
\end{longtable}

Three observations follow directly. \emph{First}, the TietAI
cross-encoder is best on every metric and at every resolution. Its
MAP@10 exact = 0.747 is +0.038 above its F1 exact = 0.709, which means
the gold code is \emph{frequently} present in the top-10 even when it
does not land at top-1; the reranker's residual error is mostly an
ordering problem, not a coverage problem. \emph{Second}, the bi-encoder
pattern is more pronounced: F1 exact = 0.359 vs MAP@10 exact = 0.461, a
+0.102 gap. The candidate retrieval is much better than top-1 suggests;
what is missing is the calibrated reranker. \emph{Third}, the three
off-the-shelf sentence-transformer baselines have similar (low) MAP@10
in the 0.22--0.29 band, with MiniLM-L6-v2 the strongest of the three on
every metric --- the larger MPNet-v2 is not.

\subsubsection{Top-1 precision, recall and
F1}\label{top-1-precision-recall-and-f1}

\begin{figure}[H]

\centering{

\includegraphics[width=0.95\linewidth,height=\textheight,keepaspectratio]{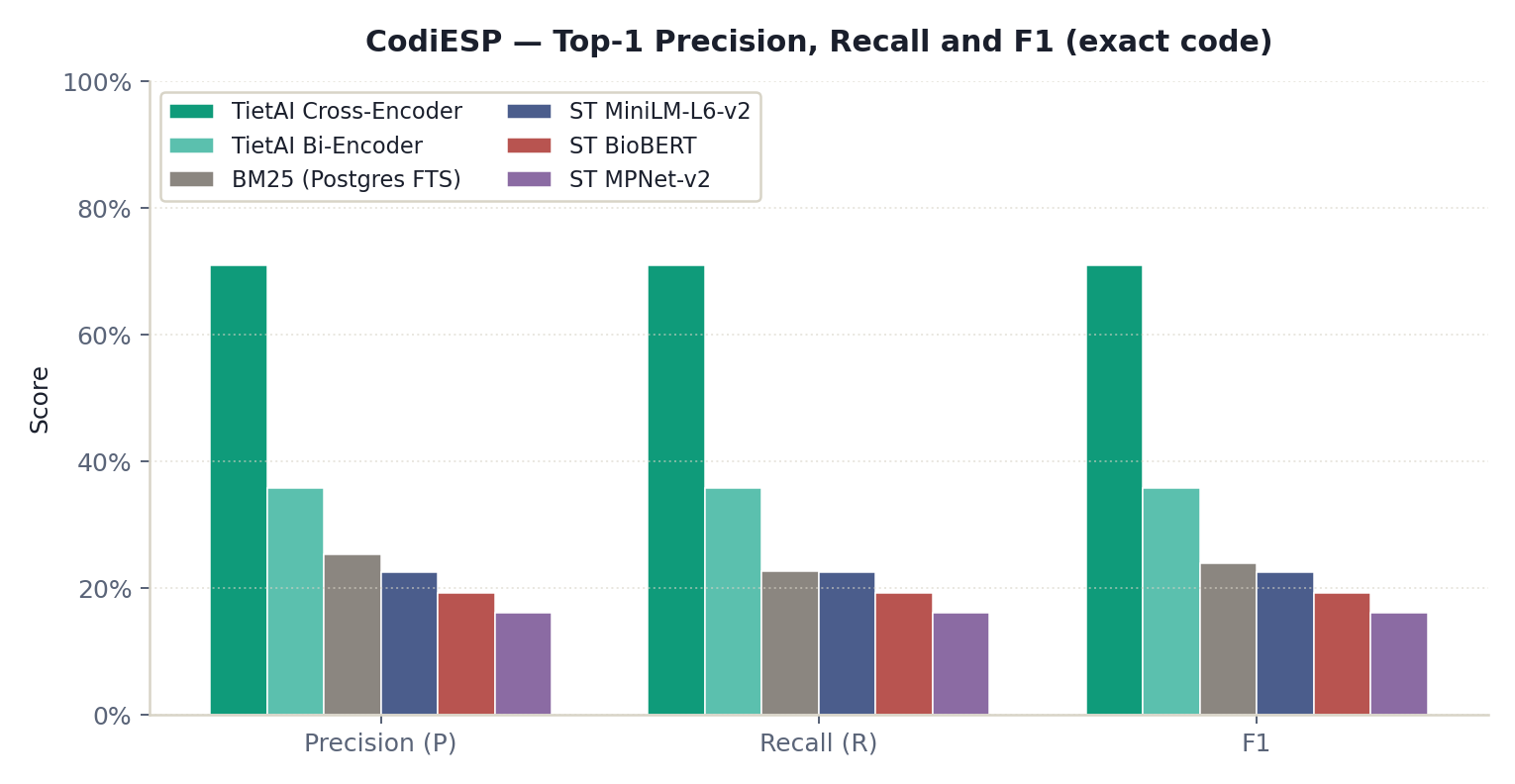}

}

\caption{\label{fig-codiesp-pr-top1}CodiESP --- Top-1 precision, recall
and F1 at the exact-code level. The TietAI Cross-Encoder dominates every
metric; the precision--recall gap visible for BM25 is the only place
these three metrics diverge.}

\end{figure}

Table~\ref{tbl-codiesp-pr-top1} and Figure~\ref{fig-codiesp-pr-top1}
isolate the precision--recall--F1 view at top-1 on the exact-code
resolution. The dense models always emit one top-1 prediction, so for
them precision = recall = F1 = accuracy. BM25 is the only model where
the three metrics diverge: it returns \emph{no} candidate for 393 of the
3,664 queries (≈11 \%), which drags its recall (0.227) below its
precision (0.254) and places its F1 (0.239) between the two. This is the
clearest empirical example of why precision alone is a misleading
summary of a coding system --- a more selective predictor can always
trade recall for precision. The full per-resolution table is in
Section~\ref{sec-annex-codiesp-full}.

\begin{longtable}[]{@{}
  >{\raggedright\arraybackslash}p{(\linewidth - 6\tabcolsep) * \real{0.4600}}
  >{\raggedleft\arraybackslash}p{(\linewidth - 6\tabcolsep) * \real{0.1800}}
  >{\raggedleft\arraybackslash}p{(\linewidth - 6\tabcolsep) * \real{0.1800}}
  >{\raggedleft\arraybackslash}p{(\linewidth - 6\tabcolsep) * \real{0.1800}}@{}}
\caption{CodiESP v4 --- Top-1 precision, recall and F1 at the exact-code
level. For dense models P = R = F1 because each query receives exactly
one prediction and has one gold code; BM25 diverges because the lexical
channel returns no candidate for \textasciitilde11 \% of
queries.}\label{tbl-codiesp-pr-top1}\tabularnewline
\toprule\noalign{}
\begin{minipage}[b]{\linewidth}\raggedright
Model
\end{minipage} & \begin{minipage}[b]{\linewidth}\raggedleft
Precision
\end{minipage} & \begin{minipage}[b]{\linewidth}\raggedleft
Recall
\end{minipage} & \begin{minipage}[b]{\linewidth}\raggedleft
F1
\end{minipage} \\
\midrule\noalign{}
\endfirsthead
\toprule\noalign{}
\begin{minipage}[b]{\linewidth}\raggedright
Model
\end{minipage} & \begin{minipage}[b]{\linewidth}\raggedleft
Precision
\end{minipage} & \begin{minipage}[b]{\linewidth}\raggedleft
Recall
\end{minipage} & \begin{minipage}[b]{\linewidth}\raggedleft
F1
\end{minipage} \\
\midrule\noalign{}
\endhead
\bottomrule\noalign{}
\endlastfoot
TietAI Cross-Encoder & \textbf{0.709} & \textbf{0.709} &
\textbf{0.709} \\
TietAI Bi-Encoder & 0.359 & 0.359 & 0.359 \\
BM25 (Postgres FTS) & 0.254 & 0.227 & 0.239 \\
ST MiniLM-L6-v2 & 0.225 & 0.225 & 0.225 \\
ST BioBERT & 0.193 & 0.193 & 0.193 \\
ST MPNet-v2 & 0.161 & 0.161 & 0.161 \\
\end{longtable}

\subsubsection{\texorpdfstring{Precision and recall across the top-\(k\)
ranking}{Precision and recall across the top-k ranking}}\label{precision-and-recall-across-the-top-k-ranking}

\begin{figure}[H]

\centering{

\includegraphics[width=1\linewidth,height=\textheight,keepaspectratio]{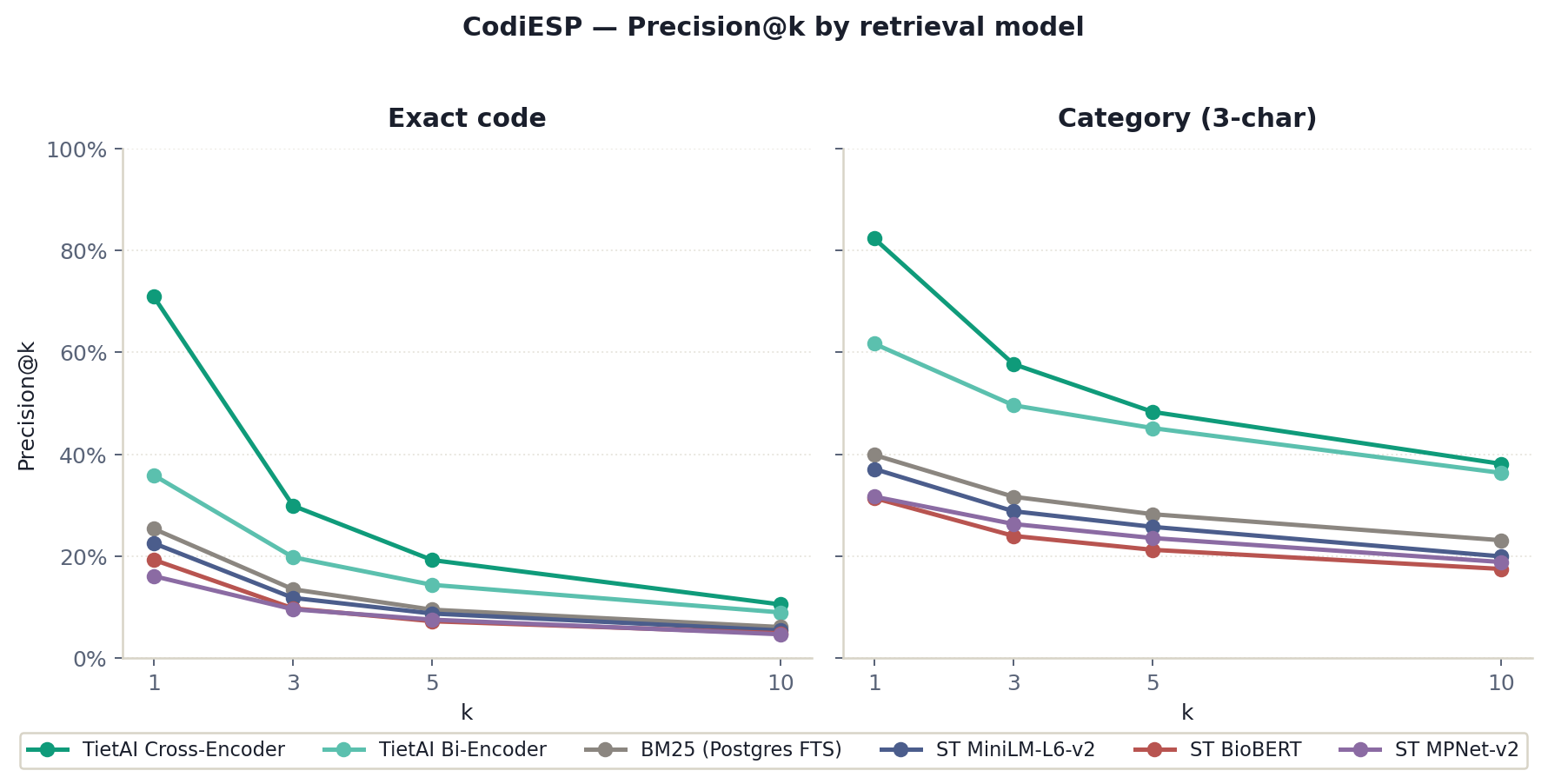}

}

\caption{\label{fig-codiesp-precision-at-k}CodiESP --- Precision@k by
retrieval model. Precision decays with k because each query has
typically one gold code, but the relative ordering of methods is
preserved at every k.}

\end{figure}

\begin{figure}[H]

\centering{

\includegraphics[width=1\linewidth,height=\textheight,keepaspectratio]{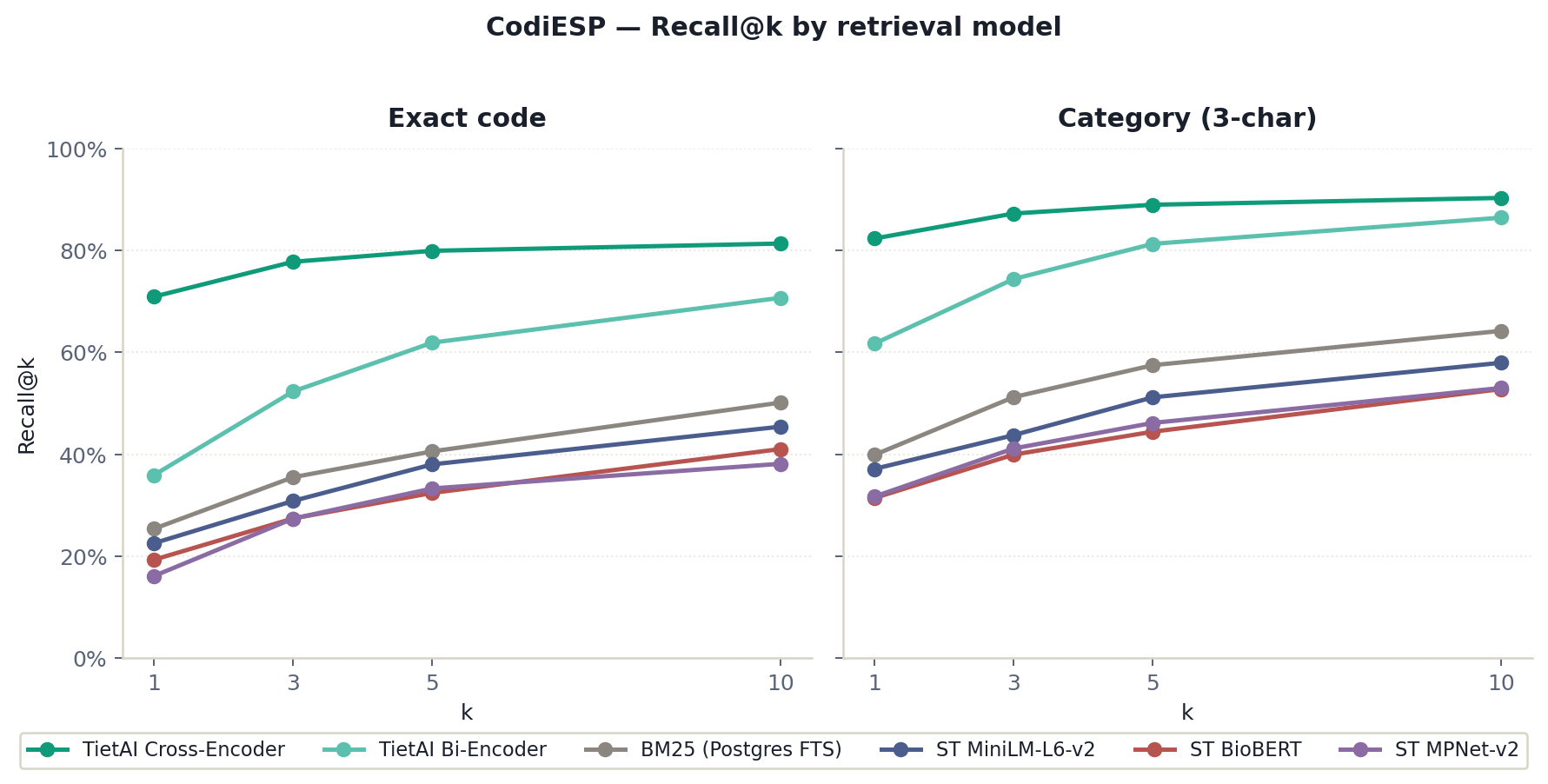}

}

\caption{\label{fig-codiesp-recall-at-k}CodiESP --- Recall@k by
retrieval model. The TietAI cross-encoder reaches R@10 ≈ 0.81 exact /
0.90 category; the bi-encoder closes most of the gap by k = 10,
confirming that the bi-encoder's recall is already strong and what the
cross-encoder supplies is the top-1 ordering.}

\end{figure}

Figure~\ref{fig-codiesp-precision-at-k} and
Figure~\ref{fig-codiesp-recall-at-k} together characterise how the
ranking quality of each retriever evolves with the cut-off \(k\).
\textbf{Precision@k decays} with \(k\) --- by construction, since each
query has typically one gold code among the top candidates, so deeper
candidates are almost always wrong --- and the cross-encoder falls from
0.709 at \(k=1\) to 0.105 at \(k=10\). \textbf{Recall@k grows} with
\(k\): the cross-encoder rises from 0.709 to 0.813 exact and from 0.823
to 0.903 at category level, and the bi-encoder rises from 0.359 to 0.707
exact and from 0.617 to 0.864 at category level. The relative ordering
of methods is preserved at every \(k\) on both axes.
Table~\ref{tbl-codiesp-recall-at-k} reports the exact-code recall@\(k\)
numbers behind the figure; the category-level breakdown is in
Section~\ref{sec-annex-codiesp-full}.

\begin{longtable}[]{@{}
  >{\raggedright\arraybackslash}p{(\linewidth - 8\tabcolsep) * \real{0.4000}}
  >{\raggedleft\arraybackslash}p{(\linewidth - 8\tabcolsep) * \real{0.1500}}
  >{\raggedleft\arraybackslash}p{(\linewidth - 8\tabcolsep) * \real{0.1500}}
  >{\raggedleft\arraybackslash}p{(\linewidth - 8\tabcolsep) * \real{0.1500}}
  >{\raggedleft\arraybackslash}p{(\linewidth - 8\tabcolsep) * \real{0.1500}}@{}}
\caption{CodiESP v4 --- Recall@k (exact-code) by retrieval model.
Defined as the fraction of queries for which the gold code appears in
the top-\(k\) retrieved
candidates.}\label{tbl-codiesp-recall-at-k}\tabularnewline
\toprule\noalign{}
\begin{minipage}[b]{\linewidth}\raggedright
Model
\end{minipage} & \begin{minipage}[b]{\linewidth}\raggedleft
R@1
\end{minipage} & \begin{minipage}[b]{\linewidth}\raggedleft
R@3
\end{minipage} & \begin{minipage}[b]{\linewidth}\raggedleft
R@5
\end{minipage} & \begin{minipage}[b]{\linewidth}\raggedleft
R@10
\end{minipage} \\
\midrule\noalign{}
\endfirsthead
\toprule\noalign{}
\begin{minipage}[b]{\linewidth}\raggedright
Model
\end{minipage} & \begin{minipage}[b]{\linewidth}\raggedleft
R@1
\end{minipage} & \begin{minipage}[b]{\linewidth}\raggedleft
R@3
\end{minipage} & \begin{minipage}[b]{\linewidth}\raggedleft
R@5
\end{minipage} & \begin{minipage}[b]{\linewidth}\raggedleft
R@10
\end{minipage} \\
\midrule\noalign{}
\endhead
\bottomrule\noalign{}
\endlastfoot
TietAI Cross-Encoder & \textbf{0.709} & \textbf{0.778} & \textbf{0.799}
& \textbf{0.813} \\
TietAI Bi-Encoder & 0.359 & 0.523 & 0.619 & 0.707 \\
BM25 (Postgres FTS) & 0.254 & 0.355 & 0.406 & 0.501 \\
ST MiniLM-L6-v2 & 0.225 & 0.308 & 0.380 & 0.454 \\
ST BioBERT & 0.193 & 0.274 & 0.324 & 0.410 \\
ST MPNet-v2 & 0.161 & 0.274 & 0.333 & 0.381 \\
\end{longtable}

\subsection{Results on DISTEMIST}\label{sec-distemist-results}

\begin{figure}[H]

\centering{

\includegraphics[width=0.8\linewidth,height=\textheight,keepaspectratio]{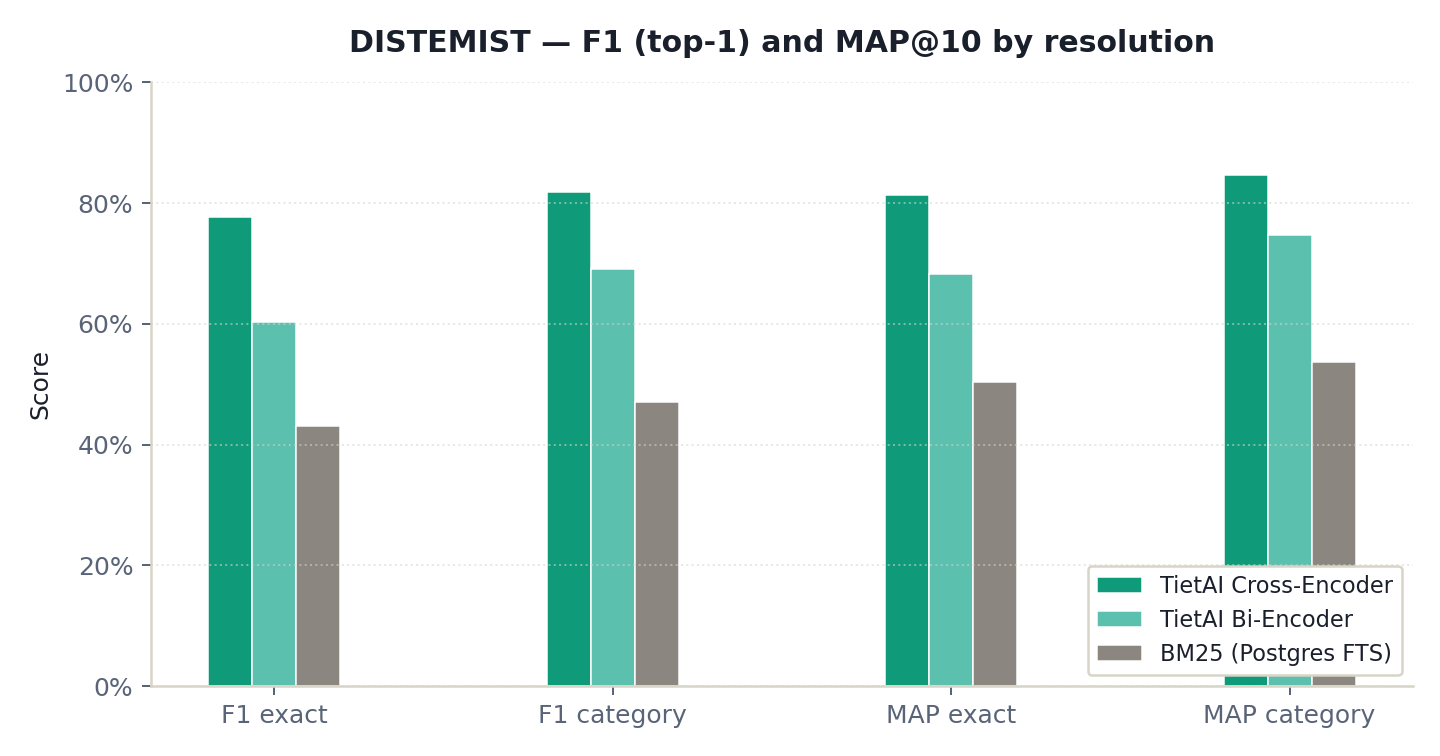}

}

\caption{\label{fig-distemist-f1-map}DISTEMIST --- F1 and MAP@10 by
model. The relative ordering of cross-encoder \textgreater{} bi-encoder
\textgreater{} BM25 holds at every metric and at both resolutions,
replicating the CodiESP pattern on a corpus the cross-encoder has not
been trained on.}

\end{figure}

We re-run the TietAI stack and the BM25 baseline on the 1,224 disease
mentions of DISTEMIST subtrack-2-linking. The aim is narrow: confirm
that the F1 and MAP rankings observed on CodiESP also hold on a corpus
with a different provenance. Table~\ref{tbl-distemist-headline} and
Figure~\ref{fig-distemist-f1-map} summarise the headline metrics; the
full breakdown is in Section~\ref{sec-annex-distemist-full}.

\begin{longtable}[]{@{}
  >{\raggedright\arraybackslash}p{(\linewidth - 8\tabcolsep) * \real{0.4000}}
  >{\raggedleft\arraybackslash}p{(\linewidth - 8\tabcolsep) * \real{0.1500}}
  >{\raggedleft\arraybackslash}p{(\linewidth - 8\tabcolsep) * \real{0.1500}}
  >{\raggedleft\arraybackslash}p{(\linewidth - 8\tabcolsep) * \real{0.1500}}
  >{\raggedleft\arraybackslash}p{(\linewidth - 8\tabcolsep) * \real{0.1500}}@{}}
\caption{DISTEMIST --- F1 at top-1 and MAP@10, at the exact-code and
three-character category
levels.}\label{tbl-distemist-headline}\tabularnewline
\toprule\noalign{}
\begin{minipage}[b]{\linewidth}\raggedright
Model
\end{minipage} & \begin{minipage}[b]{\linewidth}\raggedleft
F1 exact
\end{minipage} & \begin{minipage}[b]{\linewidth}\raggedleft
F1 category
\end{minipage} & \begin{minipage}[b]{\linewidth}\raggedleft
MAP@10 exact
\end{minipage} & \begin{minipage}[b]{\linewidth}\raggedleft
MAP@10 category
\end{minipage} \\
\midrule\noalign{}
\endfirsthead
\toprule\noalign{}
\begin{minipage}[b]{\linewidth}\raggedright
Model
\end{minipage} & \begin{minipage}[b]{\linewidth}\raggedleft
F1 exact
\end{minipage} & \begin{minipage}[b]{\linewidth}\raggedleft
F1 category
\end{minipage} & \begin{minipage}[b]{\linewidth}\raggedleft
MAP@10 exact
\end{minipage} & \begin{minipage}[b]{\linewidth}\raggedleft
MAP@10 category
\end{minipage} \\
\midrule\noalign{}
\endhead
\bottomrule\noalign{}
\endlastfoot
TietAI Cross-Encoder & \textbf{0.776} & \textbf{0.818} & \textbf{0.812}
& \textbf{0.846} \\
TietAI Bi-Encoder & 0.603 & 0.690 & 0.682 & 0.747 \\
BM25 (Postgres FTS) & 0.431 & 0.470 & 0.504 & 0.537 \\
\end{longtable}

As on CodiESP, MAP is uniformly higher than F1, by a wider margin for
the bi-encoder (+0.079 exact) and BM25 (+0.073 exact) than for the
cross-encoder (+0.036 exact). The cross-encoder's reranking is already
exploiting most of the ranking signal in the bi-encoder pool, leaving
little room for a further pass; the un-reranked stages still have
ranking quality that a downstream reranker could harvest.

BM25 again shows the precision--recall divergence at top-1 --- 86 of the
1,224 queries (≈7 \%) receive no lexical candidate, so its
\textbf{precision (0.447)} is higher than its \textbf{recall (0.416)}
and its \textbf{F1 (0.431)} lands between them (see
Section~\ref{sec-annex-distemist-full}).

\section{Discussion}\label{discussion}

The CodiESP results motivate two interpretations. First, ICD retrieval
differs from general biomedical similarity because full codes encode
qualifier-level distinctions. Second, the public-baseline ranking
suggests that Spanish language coverage matters more than English
biomedical pretraining on this benchmark.

\subsection{Why clinical-domain training does not transfer to
coding}\label{why-clinical-domain-training-does-not-transfer-to-coding}

Automatic ICD assignment aligns a free-text mention with a controlled
vocabulary entry. It is not equivalent to retrieving a broadly similar
biomedical sentence. ICD-10-CM and CIE-10-ES codes often encode disease,
anatomical site, laterality, severity, episode of care, and etiology /
manifestation structure in a single label
\citep{cms2024icd10guidelines, msssi2024cie10es}.

Code \emph{``s22.49xa''} illustrates this structure. The corresponding
description is \emph{``multiple fractures of ribs, right side, initial
encounter for closed fracture''}. The model must jointly capture
anatomy, laterality, multiplicity, fracture type, and episode
information to distinguish this code from its siblings in
\emph{``S22.4''}.

Bi-encoders trained on NLI or STS-style objectives, including
MiniLM-L6-v2 and BioBERT-ST, are optimized for sentence-level semantic
proximity. This tends to place sibling concepts such as right-sided and
left-sided rib fractures close to each other. Domain pretraining can
move representations toward biomedical content, but it does not
necessarily separate the qualifier axes required by coding guidelines.

The cross-encoder addresses this limitation by scoring each query-code
pair jointly. Its listwise hard negatives are drawn from neighbouring
CIE-10 codes, so training directly penalises loss of laterality,
episode, and severity information. This interpretation is consistent
with the CodiESP gap: the bi-encoder captures chapter-level topic
information well, but the reranker is needed for exact-code ordering.

\subsection{Why language matters more than domain on this
benchmark}\label{why-language-matters-more-than-domain-on-this-benchmark}

The public-baseline ranking also points to language coverage. BioBERT-ST
is a biomedical encoder, but its pretraining corpus is overwhelmingly
English \citep{lee2020biobert}. Spanish is widely spoken but remains a
medium-resource language in NLP, especially for biomedical corpora
\citep{joshi2020stateandfate}.

Tokenizer coverage is one plausible mechanism. English biomedical
tokenizers can over-segment Spanish medical compounds such as
\emph{broncoaspiración}, \emph{queratoconjuntivitis}, and
\emph{microhematuria}. Prior work on tokenizer fairness shows that such
over-segmentation can reduce downstream performance
\citep{rust2021howgood, ahia2023all, petrov2023language}.

Cross-lingual transfer from English biomedical models is also known to
weaken on Spanish, Catalan, and Portuguese clinical NLP tasks
\citep{neveol2018clinicalmultilang, soares2019multilingualclinical, lauscher2020zerohero}.
In our setting, both queries and candidates are Spanish. The observed
BioBERT-ST \textless{} MiniLM-L6-v2 ordering should therefore be read as
evidence against English biomedical transfer, not against biomedical
specialisation in general.

\subsection{Calibrating our numbers against the CodiESP
benchmark}\label{sec-litcomp}

The natural calibration point for our CodiESP numbers is the original
overview of the CodiESP shared task \citep{miranda2020codiesp}, whose
Table 7 reports precision, recall, F1 and MAP for the best run of each
participating team on the diagnosis sub-track (CodiEsp-D). We reproduce
the diagnosis columns in Table~\ref{tbl-codiesp2020-table7}, ranked by
F1, and add our TietAI cross-encoder result on the same benchmark
(per-mention CIE-10-ES retrieval on the 3,664 dev mentions; see caveats
below).

\begin{longtable}[]{@{}
  >{\raggedright\arraybackslash}p{(\linewidth - 8\tabcolsep) * \real{0.5000}}
  >{\raggedleft\arraybackslash}p{(\linewidth - 8\tabcolsep) * \real{0.1200}}
  >{\raggedleft\arraybackslash}p{(\linewidth - 8\tabcolsep) * \real{0.1200}}
  >{\raggedleft\arraybackslash}p{(\linewidth - 8\tabcolsep) * \real{0.1200}}
  >{\raggedleft\arraybackslash}p{(\linewidth - 8\tabcolsep) * \real{0.1400}}@{}}
\caption{CodiESP-D best-run-per-team summary, reproduced from Table 7 of
Miranda 2020 \citep{miranda2020codiesp} and augmented with our TietAI
cross-encoder (top row). Diagnosis sub-track only. Bold marks the best
value in each column.}\label{tbl-codiesp2020-table7}\tabularnewline
\toprule\noalign{}
\begin{minipage}[b]{\linewidth}\raggedright
Team / system
\end{minipage} & \begin{minipage}[b]{\linewidth}\raggedleft
Precision
\end{minipage} & \begin{minipage}[b]{\linewidth}\raggedleft
Recall
\end{minipage} & \begin{minipage}[b]{\linewidth}\raggedleft
F1
\end{minipage} & \begin{minipage}[b]{\linewidth}\raggedleft
MAP
\end{minipage} \\
\midrule\noalign{}
\endfirsthead
\toprule\noalign{}
\begin{minipage}[b]{\linewidth}\raggedright
Team / system
\end{minipage} & \begin{minipage}[b]{\linewidth}\raggedleft
Precision
\end{minipage} & \begin{minipage}[b]{\linewidth}\raggedleft
Recall
\end{minipage} & \begin{minipage}[b]{\linewidth}\raggedleft
F1
\end{minipage} & \begin{minipage}[b]{\linewidth}\raggedleft
MAP
\end{minipage} \\
\midrule\noalign{}
\endhead
\bottomrule\noalign{}
\endlastfoot
\textbf{TietAI Cross-Encoder (ours)} & \textbf{0.709} & \textbf{0.709} &
\textbf{0.709} & \textbf{0.747} \\
IAM \citep{codiesp2020iamteam} & 0.817 & 0.592 & \textbf{0.687} &
0.521 \\
FLE & 0.740 & 0.633 & 0.679 & 0.519 \\
Anuj & 0.741 & 0.621 & 0.676 & 0.505 \\
MEDIA & 0.735 & 0.630 & 0.629 & 0.488 \\
The Mental Strokers & 0.759 & 0.638 & 0.591 & 0.517 \\
UDC-UA & 0.727 & 0.605 & 0.546 & 0.368 \\
SINAI & 0.450 & 0.544 & 0.488 & 0.314 \\
IMS & 0.373 & 0.709 & 0.474 & 0.449 \\
LSI-UNED & 0.253 & 0.688 & 0.370 & 0.517 \\
BCSG & 0.457 & 0.287 & 0.337 & 0.259 \\
SWAP & 0.295 & 0.442 & 0.308 & 0.202 \\
CodeICD@IITH & 0.462 & 0.281 & 0.350 & 0.192 \\
TeamX & 0.123 & 0.858 & 0.192 & 0.299 \\
DCIC - UNS & 0.482 & 0.261 & 0.187 & 0.097 \\
NLP-UNED & 0.542 & 0.089 & 0.153 & 0.100 \\
ExeterChiefs & 0.117 & 0.201 & 0.144 & 0.082 \\
Hulat-PDPQ & \textbf{0.866} & 0.066 & 0.123 & 0.115 \\
LIIR & 0.124 & 0.055 & 0.076 & 0.044 \\
SSN-NLP & 0.025 & 0.049 & 0.033 & 0.007 \\
nlp4life & 0.014 & 0.038 & 0.020 & 0.004 \\
IXA-AAA & 0.004 & 0.858 & 0.009 & \textbf{0.593} \\
ICB-UMA & 0.004 & \textbf{0.897} & 0.009 & 0.482 \\
\end{longtable}

Two patterns dominate Table~\ref{tbl-codiesp2020-table7} and motivate
everything that follows. The benchmark is, on aggregate, dominated by
systems that have collapsed onto one of the two metrics individually.
ICB-UMA and IXA-AAA reach recall above 0.85 but at the cost of precision
below 0.005 --- they recover almost every gold code but bury it in a
flood of spurious predictions; their F1 collapses to 0.009.
Symmetrically, Hulat-PDPQ reaches precision 0.866 but recall 0.066 ---
it almost never predicts but is usually right when it does; its F1 also
collapses to 0.123. Only six teams reach an F1 above 0.5, and the top
scorer (IAM) sits at F1 = 0.687 / MAP = 0.521. Precision and recall
reported in isolation are therefore essentially uninformative on this
benchmark; F1 and MAP are the only metrics that recover a usable
ordering of the systems. This is the empirical justification for the
metric selection in Section~\ref{sec-metrics}.

Three direct conclusions follow. \emph{First}, the TietAI cross-encoder
matches or exceeds the best F1 reported in the 2020 shared task (0.709
vs 0.687) and is substantially ahead on MAP (0.747 vs 0.593).
\emph{Second}, the absolute level of the benchmark is, frankly, modest:
the best-published F1 in 2020 sits below 0.69 on diagnosis coding, and
the median best-team F1 is around 0.35. That ceiling is the operational
context any downstream coding-assistant system inherits. \emph{Third},
in the agentic-coding pipelines that have emerged since 2024 --- where
an LLM reranks a candidate shortlist \citep{cascadeicd2026} --- a
high-MAP upstream retriever is structurally much more useful than a
high-recall one. If MAP@10 is 0.75, then on three out of four mentions
the \emph{first} reranker iteration is examining a shortlist that
already contains the correct code in a near-top position; the reranker
pays for itself in roughly the number of tokens it spends on the first
prompt and avoids the long-tail iterations that high-recall /
low-precision retrievers force.

Several caveats apply to the row-level comparison. The 2020 shared task
reports \emph{document-level} metrics: precision, recall, F1 and MAP are
computed over the set of codes assigned to each clinical case. We report
\emph{per-mention} retrieval: precision, recall and F1 at top-1 are
computed over (mention, code) pairs, of which there are 3,664 in the dev
set. The two are correlated but not identical, and a system that is
perfect at per-mention coding may still be docked at document-level F1
if it duplicates codes or misses the document-level multi-label
aggregation. We also evaluate on the \emph{dev} set whereas the shared
task reports \emph{test}, and our cross-encoder has seen CodiESP-D seed
mentions during training (the cross-encoder corpus is the listwise
groups built from CodiESP, as described in
Section~\ref{sec-experimental-design}). The table should therefore be
read as an order-of-magnitude calibration, not as a head-to-head
leaderboard.

A separate observation, important for what follows in the next
subsection, is that \textbf{every system in
Table~\ref{tbl-codiesp2020-table7} was trained on CodiESP itself.} The
2020 shared task distributed training and development splits drawn from
the same 1,000-case corpus, and the test split was disjoint but from the
same distribution. Our pipeline is the only entry in the table whose
bi-encoder has never seen a CodiESP document: the contrastive bi-encoder
was trained exclusively on LLM-generated synthetic pairs from a
different, broader Spanish biomedical distribution.

\subsection{Data curation, not just model size, is the binding
constraint}\label{data-curation-not-just-model-size-is-the-binding-constraint}

The pattern in Table~\ref{tbl-codiesp2020-table7} reveals something that
the 2020 shared task itself does not stress. The participating systems
were heterogeneous --- fine-tuned multilingual BERT, classical ML
pipelines, custom rule-based codifiers --- and the absolute
top-of-leaderboard F1 sat at 0.687 even though every team had access to
the full \emph{in-distribution} training split and a generous compute
budget. Five years later, our pipeline reaches F1 = 0.709 on the same
dev mentions with a bi-encoder trained on roughly 19,500
\emph{out-of-sample} LLM-generated pairs (with the cross-encoder being
the only component that has seen CodiESP-derived listwise groups). The
architectures used in 2020 were not crude --- fine-tuned BERT is the
same family of model that backs our bi-encoder --- and yet the marginal
gain from going from a 2020 in-sample fine-tuning recipe to a 2025
out-of-sample synthetic recipe with a calibrated cross-encoder is at
least as large as the gap between the best and the median 2020 entrant.
The conclusion is arithmetically inescapable: data curation ---
specifically, generating training pairs that respect the qualifier axes
of CIE-10 (laterality, anatomy, episode, severity) and matching the
target language --- is the binding constraint on retrieval quality for
this task, not model size or the choice of backbone family.

The most striking version of this finding is the comparison of our
bi-encoder (F1 = 0.359, MAP = 0.461, trained on 19.5 k synthetic
out-of-sample pairs) with the un-tuned \texttt{roberta-base-bio-cli}
reported in ClinLinker's Table 1 \citep{gallego2024clinlinker} on
DISTEMIST (top-1 = 0.509 on SNOMED-CT). The two models share a backbone
family; the difference between them is essentially the contrastive
fine-tuning corpus and its curation strategy.

\subsection{Implications of retrieval quality concentration in the
cross-encoder}\label{implications-of-retrieval-quality-concentration-in-the-cross-encoder}

The headline tables in Section~\ref{sec-codiesp-results} and
Section~\ref{sec-distemist-results} show that the principal gains in our
pipeline come from the cross-encoder. On CodiESP exact F1, the
cross-encoder adds +0.350 over the bi-encoder; on DISTEMIST, +0.173. The
absolute levels of the bi-encoder are already useful --- its MAP@10 at
category level is 0.694 on CodiESP and 0.747 on DISTEMIST --- but the
calibration that turns top-10 retrieval into reliable top-1
recommendation is what the cross-encoder supplies. We see this in two
quantitative regularities. \emph{MAP exceeds F1 by a larger margin for
the bi-encoder than for the cross-encoder} on both corpora, which is the
signature of an under-calibrated ranker that has good recall in the
top-\(k\) pool but does not order it well. And \emph{the bi-encoder
catches up to the cross-encoder at deeper \(k\)} on DISTEMIST (within 3
pp by top-10) --- the difference is at the head of the ranked list, not
at the coverage tail.

This concentration of value in the reranking stage suggests a clean
production architecture. A \emph{single}, curated Spanish biomedical
bi-encoder can serve as the universal candidate generator across the
whole institution --- it is cheap to host, language-agnostic in
practice, and its retrieval quality is already well above any
generalistic alternative. The reranking layer, by contrast, can be
\emph{specialised}: a language-specific cross-encoder for each target
clinical language, and optionally a specialty-specific cross-encoder
(cardiology, oncology, primary care) trained on the listwise groups
relevant to that specialty's most-coded sub-trees. Specialised rerankers
are individually cheap to train (our cross-encoder is fit on ≈10,600
listwise groups) and can be applied \emph{on demand}: the front-end
router picks the right reranker per request, paying specialisation cost
only when the specialty signal is informative. The combination --- a
universal bi-encoder upstream, on-demand specialised cross-encoders
downstream --- is the configuration most consistent with the results in
this paper.

A natural extension, which we have not evaluated quantitatively but
which the metric structure invites, is to replace the cross-encoder with
an LLM-based coding agent that reads the bi-encoder's top-\(k\)
shortlist plus the document context and produces the final code
\citep{cascadeicd2026}. Because our bi-encoder already achieves
category-level MAP@10 above 0.69 (CodiESP) and 0.74 (DISTEMIST), an
agent that operates \emph{within} the category shortlist is, by
construction, working in a regime where the correct category is almost
always among its few visible options. We would expect such an agent to
approach the cross-encoder's category-level numbers (0.823 F1 on
CodiESP, 0.818 on DISTEMIST) while spending only the tokens needed for a
single category-restricted prompt --- an order-of-magnitude saving over
agents that operate on the full CIE-10-ES vocabulary or on unfiltered
BM25 lists.

\subsection{Figures beyond CIE-10-ES, ICD-coding
literature}\label{figures-beyond-cie-10-es-icd-coding-literature}

A separate calibration question is whether the F1 figures we report on
Spanish CIE-10-ES coding are within the broad envelope of published
results on full-code ICD-10 / ICD-9 coding in English. On MIMIC-III
(ICD-9 full, ≈8,900 codes), the previously dominant PLM-ICD model
\citep{plmicd2022} reports \emph{micro-F1 = 0.607} and \emph{macro-F1 =
0.290}; the more recent GoM-ICD architecture \citep{pan2025gomicd} is
the current state-of-the-art and reports \emph{micro-F1 = 0.613} on
MIMIC-IV ICD-10 full. A 2026 systematic review of agentic-LLM pipelines
on MIMIC-IV summarises a precision ceiling of about \emph{55--80 \%} for
autonomous LLM coders, with the best pipelines --- where a
domain-specific encoder produces a shortlist that an LLM then audits ---
improving precision dramatically while keeping recall in the 0.6--0.7
band \citep{cascadeicd2026, icdllmsr2025}. Our CodiESP cross-encoder (F1
= 0.709 exact / 0.823 category) and DISTEMIST cross-encoder (F1 = 0.776
exact / 0.818 category) sit at the top of that envelope and \emph{above}
the published English ICD-10 SOTA at the exact-code level --- a result
that is partly explained by our per-mention evaluation protocol (which
is easier than document-level coding) but also by the structural
advantage of a bi-encoder + cross-encoder pipeline trained explicitly on
listwise hard negatives drawn from the target vocabulary.

\section{Conclusions}\label{conclusions}

We evaluated benefits of the optimization of input data in clinical
coding models, based on semantic search (bi-encoder plus a cross encoder
reranker). We confirm that the proper data, in this case we use a LLM
frontier model as tutor, is enough to improve specialized models
trainined with large dataset: CodiESP v4 (3,664 mention-to-code pairs)
and DISTEMIST (1,224 disease mentions). On CodiESP the pipeline reaches
\emph{F1 = 0.709} at the exact-code level and \emph{MAP@10 = 0.747}; on
DISTEMIST it reaches \emph{F1 = 0.776} exact / \emph{MAP@10 = 0.812}. At
the three-character category level, the same metrics rise to \emph{0.82
/ 0.85} and \emph{0.82 / 0.85} respectively.

We also can conclude that off-the-shelf embedding models are not
appropriate for specialised clinical-coding downstream tasks, and
naïvely larger or nominally in-domain encoders make this worse, not
better. MiniLM-L6-v2 is the strongest public sentence-transformer
baseline on CodiESP, but its F1 exact (0.225) is barely above BM25 and
three times below the task-specific retriever. The 768-dimensional
MPNet-v2 is worse than the 384-dimensional MiniLM; the in-domain
S-BioBERT is worse still. The shape of these failures --- captured
cleanly by F1 and MAP rather than by precision or recall in isolation,
as the literature comparison in Section~\ref{sec-litcomp} shows --- is
that out-of-the-box encoders neither carry the right Spanish
tokenization nor have been calibrated against the qualifier-level
distinctions that CIE-10-ES encodes.

As mentioned, fine-tuning with synthetic data generated by an LLM tutor
is an effective recipe for specialised coding retrievers, and the
dataset required is surprisingly small (bi-encoder is fit on 19,502
synthetic and the cross-encoder on 10,628 listwise groups). The combined
supervised signal is several orders of magnitude smaller than the
corpora used to pre-train BioBERT (≈4.5 B word tokens) or to train
modern multilingual SapBERT variants (over a million UMLS Spanish
concepts). The 2020 CodiESP shared-task entrants used in-distribution
training data drawn from the same corpus as the test set and reached at
most F1 = 0.687 (see Table~\ref{tbl-codiesp2020-table7}). Our pipeline
reaches F1 = 0.709 on the same dev mentions with a bi-encoder that has
never seen a CodiESP document and a cross-encoder trained on a listwise
corpus an order of magnitude smaller than what most participating teams
had. Data curation --- the alignment of the training distribution with
the qualifier axes of the target vocabulary and the target language ---
is the binding constraint, not the volume of pre-training tokens.

Nevertyheless, retrieval quality concentrates in the cross-encoder
reranking stage, which has direct architectural implications for
production deployment. On CodiESP, the cross-encoder adds +0.350 to
exact F1 over the bi-encoder on the same candidate pool; on DISTEMIST,
+0.173. The bi-encoder's MAP@10 is already in the 0.69--0.75 band at
category level, which means the right candidate is almost always in the
shortlist; what the cross-encoder supplies is the calibrated ordering.
The clean production architecture is therefore a \emph{universal curated
bi-encoder} shared across all clinical workloads, plus
\emph{language-specific (and optionally specialty-specific)
cross-encoders} applied on demand. The same shortlist could equally well
be consumed by an LLM-based coding agent, in particular we believe an
agenitc reranker on top of a two-step encoder would yield the optimal
results.

\section{Future work}\label{future-work}

Three directions are needed to firm up the language-coverage claim and
to push absolute numbers further.

\begin{enumerate}
\def\labelenumi{\arabic{enumi}.}
\item
  \emph{A multilingual evaluation that isolates the language effect.}
  All results in this paper are reported on Spanish corpora. To confirm
  that the BioBERT-ST regression is driven by language coverage rather
  than by domain specialisation, the next experiment is to re-run the
  same benchmark on English (where BioBERT should regain its expected
  advantage) and on languages syntactically and lexically close to
  Spanish --- Italian and Portuguese in particular, and Catalan as a
  stress test. A roughly flat ranking across romance languages and a
  sharp re-ordering on English would be the diagnostic signature of the
  language-coverage hypothesis.
\item
  \emph{Training with a substantially larger synthetic corpus.} The 19.5
  k synthetic pairs used here were sufficient to outperform every public
  baseline and the 2020 CodiESP leaderboard at the diagnosis level, but
  the learning curves had not yet plateaued at the cut-off. Scaling
  Dataset A by an order of magnitude --- by extending the LLM data
  factory to all 22 ICD-10 chapters in all six languages, and by mining
  additional positives from existing CIE-10-ES indexed corpora --- is
  the obvious next step. The marginal gain should concentrate on rare
  qualifier combinations (laterality × episode × severity), which are
  the codes on which the cross-encoder still confuses siblings.
\item
  \emph{Replacing the cross-encoder with an LLM coding agent.} The
  production architecture sketched above invites a quantitative
  follow-up: how close can a small LLM, operating on the bi-encoder's
  top-\(k\) category shortlist, get to the cross-encoder's
  category-level F1, and at what token cost? Recent work on agentic
  ICD-10 pipelines \citep{cascadeicd2026, icdllmsr2025} suggests that
  the answer depends crucially on the quality of the upstream shortlist
  --- which is precisely what the LLM-tutor-trained bi-encoder is shown
  to deliver in this paper.
\end{enumerate}

\section*{Acknowledgements and Conflicts of
Interest}\label{acknowledgements-and-conflicts-of-interest}
\addcontentsline{toc}{section}{Acknowledgements and Conflicts of
Interest}

The authors thank the TietAI clinical team for their feedback and help
in the elaboration of this work.

\bibliographystyle{unsrt}
\nocite{*}
\bibliography{refs}
\clearpage

\appendix

\section{Annex I. Representative training dataset
rows}\label{sec-annex-dataset-examples}

A representative bi-encoder row (Catalan):

\begin{Shaded}
\begin{Highlighting}[]
\FunctionTok{\{}
  \DataTypeTok{"query"}\FunctionTok{:} \StringTok{"Diabetis mellitus tipus 2 no controlada"}\FunctionTok{,}
  \DataTypeTok{"positives"}\FunctionTok{:} \OtherTok{[}\StringTok{"Tinc el sucre molt alt"}\OtherTok{,}
                \StringTok{"La diabetis d\textquotesingle{}adult està descompensada"}\OtherTok{]}\FunctionTok{,}
  \DataTypeTok{"hard\_negatives"}\FunctionTok{:} \OtherTok{[}\StringTok{"La tiroide em funciona poc"}\OtherTok{,}
                     \StringTok{"Falta de vitamina D als ossos"}\OtherTok{]}\FunctionTok{,}
  \DataTypeTok{"soft\_negatives"}\FunctionTok{:} \OtherTok{[}\StringTok{"Infecció d\textquotesingle{}orina"}\OtherTok{,}
                     \StringTok{"Mal de queixal fort"}\OtherTok{]}\FunctionTok{,}
  \DataTypeTok{"chapter"}\FunctionTok{:} \StringTok{"Endocrine, nutritional and metabolic diseases (E)"}
\FunctionTok{\}}
\end{Highlighting}
\end{Shaded}

A representative cross-encoder row (Spanish, CodiEsp-derived):

\begin{Shaded}
\begin{Highlighting}[]
\FunctionTok{\{}
  \DataTypeTok{"query"}\FunctionTok{:} \StringTok{"exposición a Brucella"}\FunctionTok{,}
  \DataTypeTok{"positives"}\FunctionTok{:} \OtherTok{[}
    \StringTok{"Contacto y sospecha de exposición a enfermedad bacteriana"}
  \OtherTok{]}\FunctionTok{,}
  \DataTypeTok{"negatives"}\FunctionTok{:} \OtherTok{[}
    \StringTok{"Brucelosis debida a Brucella suis"}\OtherTok{,}
    \StringTok{"Brucelosis debida a Brucella canis"}\OtherTok{,}
    \StringTok{"Brucelosis, no especificada"}\OtherTok{,}
    \StringTok{"..."}
  \OtherTok{]}
\FunctionTok{\}}
\end{Highlighting}
\end{Shaded}

\section{Annex II. Representative evaluation dataset
rows}\label{sec-annex-eval-examples}

\subsection{CodiESP v4}\label{codiesp-v4}

CodiESP distributes one row per (document, code) annotation. Diagnoses
live in \texttt{testD.tsv}, procedures in \texttt{testP.tsv}, and the
text spans they were extracted from in \texttt{testX.tsv}. The columns
are
\texttt{\textless{}doc\_id\textgreater{}\textbackslash{}t\textless{}code\textgreater{}}
for the gold files and
\texttt{\textless{}doc\_id\textgreater{}\textbackslash{}t\textless{}label\textgreater{}\textbackslash{}t\textless{}code\textgreater{}\textbackslash{}t\textless{}span\textgreater{}\textbackslash{}t\textless{}offsets\textgreater{}}
for the spans file. Long document identifiers are abbreviated below for
page width:

\begin{Shaded}
\begin{Highlighting}[]
\NormalTok{\# testD.tsv  (CIE{-}10{-}ES diagnoses)}
\NormalTok{S0004{-}...{-}1   s22.49xa}
\NormalTok{S0004{-}...{-}1   n28.1}
\NormalTok{S0004{-}...{-}1   r69}
\NormalTok{S0004{-}...{-}1   f17.210}
\NormalTok{S0004{-}...{-}1   r31.9}

\NormalTok{\# testP.tsv  (CIE{-}10{-}ES procedures)}
\NormalTok{S0004{-}...{-}1   0ttb}
\NormalTok{S0004{-}...{-}1   bv49zzz}
\NormalTok{S0004{-}...{-}1   0djdxzz}

\NormalTok{\# testX.tsv  (mention spans, used as queries)}
\NormalTok{S0004{-}...{-}1   DIAG  s22.49xa  costales                  182 190}
\NormalTok{S0004{-}...{-}1   PROC  0ttb      resección vejiga          1847 1856}
\NormalTok{S0004{-}...{-}1   DIAG  n28.1     quistes riñón derecho     1260 1303}
\NormalTok{S0004{-}...{-}1   DIAG  r31.0     hematuria macroscópica    444 466}
\end{Highlighting}
\end{Shaded}

At evaluation time we use the span text (\texttt{costales},
\texttt{resección\ vejiga},
\texttt{quistes\ corticales\ simples\ en\ riñón\ derecho}, \ldots) as
the query and the adjacent code as the gold label.

\subsection{DISTEMIST}\label{distemist}

DISTEMIST distributes one row per disease mention with cross-mappings to
several controlled vocabularies. We use the \texttt{icd10\_code} column
as the gold CIE-10 label and the \texttt{span} column as the query:

\begin{Shaded}
\begin{Highlighting}[]
\NormalTok{filename          mark  label       span                         icd10\_code}
\NormalTok{es{-}S1139{-}...{-}1    T1    ENFERMEDAD  apéndice cecal inflamatorio   K37}
\NormalTok{es{-}S1139{-}...{-}1    T3    ENFERMEDAD  abdomen agudo                 R10.0}
\NormalTok{es{-}S1139{-}...{-}1    T4    ENFERMEDAD  apendicitis aguda             K35.9}
\NormalTok{es{-}S0210{-}...{-}3    T1    ENFERMEDAD  DM                            E14}
\NormalTok{es{-}S0210{-}...{-}3    T4    ENFERMEDAD  pancreatitis aguda            K85}
\end{Highlighting}
\end{Shaded}

Rows without an \texttt{icd10\_code} value are discarded; the remaining
1,224 (query, gold-code) pairs are the evaluation set reported in
Table~\ref{tbl-distemist-headline} and
Section~\ref{sec-annex-distemist-full}.

\clearpage

\section{Annex III. Full CodiESP per-model
metrics}\label{sec-annex-codiesp-full}

Table~\ref{tbl-annex-codiesp-top1} reports the full top-1 breakdown for
every retriever on the 3,664 CodiESP dev mentions, at the three
CIE-10-ES resolutions. For the dense models, precision = recall =
accuracy = F1 because each query receives exactly one top-1 prediction
and has one gold code. BM25 is the only model where the four metrics
diverge: it returns \emph{no} candidate for 393 of the 3,664 queries
(≈11 \%), which depresses recall and accuracy below precision.

\begin{longtable}[]{@{}
  >{\raggedright\arraybackslash}p{(\linewidth - 10\tabcolsep) * \real{0.3000}}
  >{\raggedright\arraybackslash}p{(\linewidth - 10\tabcolsep) * \real{0.1800}}
  >{\raggedleft\arraybackslash}p{(\linewidth - 10\tabcolsep) * \real{0.1300}}
  >{\raggedleft\arraybackslash}p{(\linewidth - 10\tabcolsep) * \real{0.1300}}
  >{\raggedleft\arraybackslash}p{(\linewidth - 10\tabcolsep) * \real{0.1300}}
  >{\raggedleft\arraybackslash}p{(\linewidth - 10\tabcolsep) * \real{0.1300}}@{}}
\caption{CodiESP v4 --- Full top-1 metric breakdown by model and
CIE-10-ES resolution.}\label{tbl-annex-codiesp-top1}\tabularnewline
\toprule\noalign{}
\begin{minipage}[b]{\linewidth}\raggedright
Model
\end{minipage} & \begin{minipage}[b]{\linewidth}\raggedright
Level
\end{minipage} & \begin{minipage}[b]{\linewidth}\raggedleft
P
\end{minipage} & \begin{minipage}[b]{\linewidth}\raggedleft
R
\end{minipage} & \begin{minipage}[b]{\linewidth}\raggedleft
Acc
\end{minipage} & \begin{minipage}[b]{\linewidth}\raggedleft
F1
\end{minipage} \\
\midrule\noalign{}
\endfirsthead
\toprule\noalign{}
\begin{minipage}[b]{\linewidth}\raggedright
Model
\end{minipage} & \begin{minipage}[b]{\linewidth}\raggedright
Level
\end{minipage} & \begin{minipage}[b]{\linewidth}\raggedleft
P
\end{minipage} & \begin{minipage}[b]{\linewidth}\raggedleft
R
\end{minipage} & \begin{minipage}[b]{\linewidth}\raggedleft
Acc
\end{minipage} & \begin{minipage}[b]{\linewidth}\raggedleft
F1
\end{minipage} \\
\midrule\noalign{}
\endhead
\bottomrule\noalign{}
\endlastfoot
TietAI Cross-Encoder & exact & 0.709 & 0.709 & 0.709 & 0.709 \\
TietAI Cross-Encoder & category & 0.823 & 0.823 & 0.823 & 0.823 \\
TietAI Cross-Encoder & chapter & 0.909 & 0.909 & 0.909 & 0.909 \\
TietAI Bi-Encoder & exact & 0.359 & 0.359 & 0.359 & 0.359 \\
TietAI Bi-Encoder & category & 0.617 & 0.617 & 0.617 & 0.617 \\
TietAI Bi-Encoder & chapter & 0.778 & 0.778 & 0.778 & 0.778 \\
BM25 (Postgres FTS) & exact & 0.254 & 0.227 & 0.227 & 0.239 \\
BM25 (Postgres FTS) & category & 0.399 & 0.356 & 0.356 & 0.376 \\
BM25 (Postgres FTS) & chapter & 0.559 & 0.499 & 0.499 & 0.527 \\
ST MiniLM-L6-v2 & exact & 0.225 & 0.225 & 0.225 & 0.225 \\
ST MiniLM-L6-v2 & category & 0.371 & 0.371 & 0.371 & 0.371 \\
ST MiniLM-L6-v2 & chapter & 0.521 & 0.521 & 0.521 & 0.521 \\
ST BioBERT & exact & 0.193 & 0.193 & 0.193 & 0.193 \\
ST BioBERT & category & 0.314 & 0.314 & 0.314 & 0.314 \\
ST BioBERT & chapter & 0.467 & 0.467 & 0.467 & 0.467 \\
ST MPNet-v2 & exact & 0.161 & 0.161 & 0.161 & 0.161 \\
ST MPNet-v2 & category & 0.317 & 0.317 & 0.317 & 0.317 \\
ST MPNet-v2 & chapter & 0.454 & 0.454 & 0.454 & 0.454 \\
\end{longtable}

Table~\ref{tbl-annex-codiesp-topk} reports precision@\(k\) for
\(k \in \{1,3,5,10\}\) together with MAP@10, at the exact-code and
category levels.

\begin{longtable}[]{@{}
  >{\raggedright\arraybackslash}p{(\linewidth - 12\tabcolsep) * \real{0.3000}}
  >{\raggedright\arraybackslash}p{(\linewidth - 12\tabcolsep) * \real{0.1800}}
  >{\raggedleft\arraybackslash}p{(\linewidth - 12\tabcolsep) * \real{0.0900}}
  >{\raggedleft\arraybackslash}p{(\linewidth - 12\tabcolsep) * \real{0.0900}}
  >{\raggedleft\arraybackslash}p{(\linewidth - 12\tabcolsep) * \real{0.0900}}
  >{\raggedleft\arraybackslash}p{(\linewidth - 12\tabcolsep) * \real{0.0900}}
  >{\raggedleft\arraybackslash}p{(\linewidth - 12\tabcolsep) * \real{0.1600}}@{}}
\caption{CodiESP v4 --- Precision@k (k = 1, 3, 5, 10) and MAP@10 by
model, at the exact-code and three-character category levels. Precision
drops with k because each query has typically one gold code; MAP@10
captures the ranking
quality.}\label{tbl-annex-codiesp-topk}\tabularnewline
\toprule\noalign{}
\begin{minipage}[b]{\linewidth}\raggedright
Model
\end{minipage} & \begin{minipage}[b]{\linewidth}\raggedright
Level
\end{minipage} & \begin{minipage}[b]{\linewidth}\raggedleft
P@1
\end{minipage} & \begin{minipage}[b]{\linewidth}\raggedleft
P@3
\end{minipage} & \begin{minipage}[b]{\linewidth}\raggedleft
P@5
\end{minipage} & \begin{minipage}[b]{\linewidth}\raggedleft
P@10
\end{minipage} & \begin{minipage}[b]{\linewidth}\raggedleft
MAP@10
\end{minipage} \\
\midrule\noalign{}
\endfirsthead
\toprule\noalign{}
\begin{minipage}[b]{\linewidth}\raggedright
Model
\end{minipage} & \begin{minipage}[b]{\linewidth}\raggedright
Level
\end{minipage} & \begin{minipage}[b]{\linewidth}\raggedleft
P@1
\end{minipage} & \begin{minipage}[b]{\linewidth}\raggedleft
P@3
\end{minipage} & \begin{minipage}[b]{\linewidth}\raggedleft
P@5
\end{minipage} & \begin{minipage}[b]{\linewidth}\raggedleft
P@10
\end{minipage} & \begin{minipage}[b]{\linewidth}\raggedleft
MAP@10
\end{minipage} \\
\midrule\noalign{}
\endhead
\bottomrule\noalign{}
\endlastfoot
TietAI Cross-Encoder & exact & 0.709 & 0.299 & 0.192 & 0.105 & 0.747 \\
TietAI Cross-Encoder & category & 0.823 & 0.577 & 0.483 & 0.381 &
0.851 \\
TietAI Bi-Encoder & exact & 0.359 & 0.198 & 0.144 & 0.090 & 0.461 \\
TietAI Bi-Encoder & category & 0.617 & 0.496 & 0.451 & 0.363 & 0.694 \\
BM25 (Postgres FTS) & exact & 0.254 & 0.135 & 0.095 & 0.061 & 0.322 \\
BM25 (Postgres FTS) & category & 0.399 & 0.317 & 0.282 & 0.231 &
0.471 \\
ST MiniLM-L6-v2 & exact & 0.225 & 0.118 & 0.088 & 0.054 & 0.287 \\
ST MiniLM-L6-v2 & category & 0.371 & 0.288 & 0.258 & 0.200 & 0.426 \\
ST BioBERT & exact & 0.193 & 0.098 & 0.072 & 0.049 & 0.252 \\
ST BioBERT & category & 0.314 & 0.240 & 0.213 & 0.175 & 0.373 \\
ST MPNet-v2 & exact & 0.161 & 0.096 & 0.076 & 0.047 & 0.226 \\
ST MPNet-v2 & category & 0.317 & 0.263 & 0.236 & 0.189 & 0.376 \\
\end{longtable}

Table~\ref{tbl-annex-codiesp-recall} reports the companion Recall@\(k\)
figures (fraction of queries for which the gold code appears in the
top-\(k\)), which is the metric plotted in
Figure~\ref{fig-codiesp-recall-at-k}.

\begin{longtable}[]{@{}
  >{\raggedright\arraybackslash}p{(\linewidth - 10\tabcolsep) * \real{0.3000}}
  >{\raggedright\arraybackslash}p{(\linewidth - 10\tabcolsep) * \real{0.1800}}
  >{\raggedleft\arraybackslash}p{(\linewidth - 10\tabcolsep) * \real{0.1300}}
  >{\raggedleft\arraybackslash}p{(\linewidth - 10\tabcolsep) * \real{0.1300}}
  >{\raggedleft\arraybackslash}p{(\linewidth - 10\tabcolsep) * \real{0.1300}}
  >{\raggedleft\arraybackslash}p{(\linewidth - 10\tabcolsep) * \real{0.1300}}@{}}
\caption{CodiESP v4 --- Recall@k (k = 1, 3, 5, 10) by retrieval model,
at the exact-code and three-character category levels. Recall@k here
denotes the fraction of queries for which the gold code appears in the
top-\(k\) retrieved
candidates.}\label{tbl-annex-codiesp-recall}\tabularnewline
\toprule\noalign{}
\begin{minipage}[b]{\linewidth}\raggedright
Model
\end{minipage} & \begin{minipage}[b]{\linewidth}\raggedright
Level
\end{minipage} & \begin{minipage}[b]{\linewidth}\raggedleft
R@1
\end{minipage} & \begin{minipage}[b]{\linewidth}\raggedleft
R@3
\end{minipage} & \begin{minipage}[b]{\linewidth}\raggedleft
R@5
\end{minipage} & \begin{minipage}[b]{\linewidth}\raggedleft
R@10
\end{minipage} \\
\midrule\noalign{}
\endfirsthead
\toprule\noalign{}
\begin{minipage}[b]{\linewidth}\raggedright
Model
\end{minipage} & \begin{minipage}[b]{\linewidth}\raggedright
Level
\end{minipage} & \begin{minipage}[b]{\linewidth}\raggedleft
R@1
\end{minipage} & \begin{minipage}[b]{\linewidth}\raggedleft
R@3
\end{minipage} & \begin{minipage}[b]{\linewidth}\raggedleft
R@5
\end{minipage} & \begin{minipage}[b]{\linewidth}\raggedleft
R@10
\end{minipage} \\
\midrule\noalign{}
\endhead
\bottomrule\noalign{}
\endlastfoot
TietAI Cross-Encoder & exact & 0.709 & 0.778 & 0.799 & 0.813 \\
TietAI Cross-Encoder & category & 0.823 & 0.872 & 0.889 & 0.903 \\
TietAI Bi-Encoder & exact & 0.359 & 0.523 & 0.619 & 0.707 \\
TietAI Bi-Encoder & category & 0.617 & 0.744 & 0.813 & 0.864 \\
BM25 (Postgres FTS) & exact & 0.254 & 0.355 & 0.406 & 0.501 \\
BM25 (Postgres FTS) & category & 0.399 & 0.512 & 0.575 & 0.642 \\
ST MiniLM-L6-v2 & exact & 0.225 & 0.308 & 0.380 & 0.454 \\
ST MiniLM-L6-v2 & category & 0.371 & 0.438 & 0.512 & 0.579 \\
ST BioBERT & exact & 0.193 & 0.274 & 0.324 & 0.410 \\
ST BioBERT & category & 0.314 & 0.399 & 0.444 & 0.528 \\
ST MPNet-v2 & exact & 0.161 & 0.274 & 0.333 & 0.381 \\
ST MPNet-v2 & category & 0.317 & 0.411 & 0.462 & 0.530 \\
\end{longtable}

\clearpage

\section{Annex IV. Full DISTEMIST per-model
metrics}\label{sec-annex-distemist-full}

Table~\ref{tbl-annex-distemist-top1} reports the same top-1 breakdown
for DISTEMIST. The pattern matches CodiESP: dense models collapse P = R
= Acc = F1; BM25 is the only system with divergence because 86 of the
1,224 queries (≈7 \%) receive no lexical candidate.

\begin{longtable}[]{@{}
  >{\raggedright\arraybackslash}p{(\linewidth - 10\tabcolsep) * \real{0.3000}}
  >{\raggedright\arraybackslash}p{(\linewidth - 10\tabcolsep) * \real{0.1800}}
  >{\raggedleft\arraybackslash}p{(\linewidth - 10\tabcolsep) * \real{0.1300}}
  >{\raggedleft\arraybackslash}p{(\linewidth - 10\tabcolsep) * \real{0.1300}}
  >{\raggedleft\arraybackslash}p{(\linewidth - 10\tabcolsep) * \real{0.1300}}
  >{\raggedleft\arraybackslash}p{(\linewidth - 10\tabcolsep) * \real{0.1300}}@{}}
\caption{DISTEMIST --- Full top-1 metric breakdown by model and
CIE-10-ES resolution.}\label{tbl-annex-distemist-top1}\tabularnewline
\toprule\noalign{}
\begin{minipage}[b]{\linewidth}\raggedright
Model
\end{minipage} & \begin{minipage}[b]{\linewidth}\raggedright
Level
\end{minipage} & \begin{minipage}[b]{\linewidth}\raggedleft
P
\end{minipage} & \begin{minipage}[b]{\linewidth}\raggedleft
R
\end{minipage} & \begin{minipage}[b]{\linewidth}\raggedleft
Acc
\end{minipage} & \begin{minipage}[b]{\linewidth}\raggedleft
F1
\end{minipage} \\
\midrule\noalign{}
\endfirsthead
\toprule\noalign{}
\begin{minipage}[b]{\linewidth}\raggedright
Model
\end{minipage} & \begin{minipage}[b]{\linewidth}\raggedright
Level
\end{minipage} & \begin{minipage}[b]{\linewidth}\raggedleft
P
\end{minipage} & \begin{minipage}[b]{\linewidth}\raggedleft
R
\end{minipage} & \begin{minipage}[b]{\linewidth}\raggedleft
Acc
\end{minipage} & \begin{minipage}[b]{\linewidth}\raggedleft
F1
\end{minipage} \\
\midrule\noalign{}
\endhead
\bottomrule\noalign{}
\endlastfoot
TietAI Cross-Encoder & exact & 0.776 & 0.776 & 0.776 & 0.776 \\
TietAI Cross-Encoder & category & 0.818 & 0.818 & 0.818 & 0.818 \\
TietAI Cross-Encoder & chapter & 0.926 & 0.926 & 0.926 & 0.926 \\
TietAI Bi-Encoder & exact & 0.603 & 0.603 & 0.603 & 0.603 \\
TietAI Bi-Encoder & category & 0.690 & 0.690 & 0.690 & 0.690 \\
TietAI Bi-Encoder & chapter & 0.864 & 0.864 & 0.864 & 0.864 \\
BM25 (Postgres FTS) & exact & 0.447 & 0.416 & 0.416 & 0.431 \\
BM25 (Postgres FTS) & category & 0.488 & 0.453 & 0.453 & 0.470 \\
BM25 (Postgres FTS) & chapter & 0.670 & 0.623 & 0.623 & 0.645 \\
\end{longtable}

\begin{longtable}[]{@{}
  >{\raggedright\arraybackslash}p{(\linewidth - 12\tabcolsep) * \real{0.3000}}
  >{\raggedright\arraybackslash}p{(\linewidth - 12\tabcolsep) * \real{0.1800}}
  >{\raggedleft\arraybackslash}p{(\linewidth - 12\tabcolsep) * \real{0.0900}}
  >{\raggedleft\arraybackslash}p{(\linewidth - 12\tabcolsep) * \real{0.0900}}
  >{\raggedleft\arraybackslash}p{(\linewidth - 12\tabcolsep) * \real{0.0900}}
  >{\raggedleft\arraybackslash}p{(\linewidth - 12\tabcolsep) * \real{0.0900}}
  >{\raggedleft\arraybackslash}p{(\linewidth - 12\tabcolsep) * \real{0.1600}}@{}}
\caption{DISTEMIST --- Precision@k (k = 1, 3, 5, 10) and MAP@10 by
model, at the exact-code and three-character category
levels.}\label{tbl-annex-distemist-topk}\tabularnewline
\toprule\noalign{}
\begin{minipage}[b]{\linewidth}\raggedright
Model
\end{minipage} & \begin{minipage}[b]{\linewidth}\raggedright
Level
\end{minipage} & \begin{minipage}[b]{\linewidth}\raggedleft
P@1
\end{minipage} & \begin{minipage}[b]{\linewidth}\raggedleft
P@3
\end{minipage} & \begin{minipage}[b]{\linewidth}\raggedleft
P@5
\end{minipage} & \begin{minipage}[b]{\linewidth}\raggedleft
P@10
\end{minipage} & \begin{minipage}[b]{\linewidth}\raggedleft
MAP@10
\end{minipage} \\
\midrule\noalign{}
\endfirsthead
\toprule\noalign{}
\begin{minipage}[b]{\linewidth}\raggedright
Model
\end{minipage} & \begin{minipage}[b]{\linewidth}\raggedright
Level
\end{minipage} & \begin{minipage}[b]{\linewidth}\raggedleft
P@1
\end{minipage} & \begin{minipage}[b]{\linewidth}\raggedleft
P@3
\end{minipage} & \begin{minipage}[b]{\linewidth}\raggedleft
P@5
\end{minipage} & \begin{minipage}[b]{\linewidth}\raggedleft
P@10
\end{minipage} & \begin{minipage}[b]{\linewidth}\raggedleft
MAP@10
\end{minipage} \\
\midrule\noalign{}
\endhead
\bottomrule\noalign{}
\endlastfoot
TietAI Cross-Encoder & exact & 0.776 & 0.490 & 0.408 & 0.308 & 0.812 \\
TietAI Cross-Encoder & category & 0.818 & 0.559 & 0.481 & 0.386 &
0.846 \\
TietAI Bi-Encoder & exact & 0.603 & 0.478 & 0.425 & 0.345 & 0.682 \\
TietAI Bi-Encoder & category & 0.690 & 0.537 & 0.483 & 0.398 & 0.747 \\
BM25 (Postgres FTS) & exact & 0.447 & 0.335 & 0.312 & 0.267 & 0.504 \\
BM25 (Postgres FTS) & category & 0.488 & 0.375 & 0.350 & 0.292 &
0.537 \\
\end{longtable}

\end{document}